\pdfoutput=1

\documentclass[11pt]{article}
\usepackage[table]{xcolor} 

\usepackage[final]{ACL2023}

\usepackage{booktabs}
\usepackage{multirow}
\usepackage{geometry}
\usepackage{makecell}

\usepackage{amsmath}
\usepackage{graphicx}
\usepackage{xparse}

\usepackage{times}
\usepackage{latexsym}

\usepackage{booktabs, multirow} 
\usepackage{soul}
\usepackage{changepage,threeparttable} 

\usepackage{graphicx}

\usepackage{pifont}

\usepackage{array}
\usepackage{xcolor}
\usepackage{tcolorbox}

\newcommand{\cmark}{\textcolor{green}{\ding{51}}}

\usepackage[T1]{fontenc}

\usepackage[utf8]{inputenc}

\usepackage{microtype}

\usepackage{inconsolata}

\title{Enhancing Temporal Understanding in LLMs for Semi-structured Tables}

\author{Irwin Deng,  Kushagra Dixit\thanks{~~Work done during internship at UPenn.~}~,  Vivek Gupta\thanks{~~corresponding author~}~,  Dan Roth \\
        University of Pennsylvania \\
        \tt \small { \{ideng, dixitk, gvivek, danroth\}@seas.upenn.edu}}

\begin{document}
\maketitle

\begin{abstract}

Temporal reasoning over tabular data presents substantial challenges for large language models (LLMs), as evidenced by recent research. In this study, we conduct a comprehensive analysis of temporal datasets to pinpoint the specific limitations of LLMs. Our investigation leads to enhancements in TempTabQA, a dataset specifically designed for tabular temporal question answering. We provide critical insights for improving LLM performance in temporal reasoning tasks with tabular data. Furthermore, we introduce a novel approach, C.L.E.A.R to strengthen LLM capabilities in this domain. Our findings demonstrate that our method significantly improves evidence-based reasoning across various models. Additionally, our experimental results reveal that indirect supervision with auxiliary data substantially boosts model performance in these tasks. This work contributes to a deeper understanding of LLMs' temporal reasoning abilities over tabular data and promotes advancements in their application across diverse fields.

\end{abstract}

\section{Introduction}

\begin{figure}[!ht]
\centering
  \includegraphics[width=0.7\columnwidth]{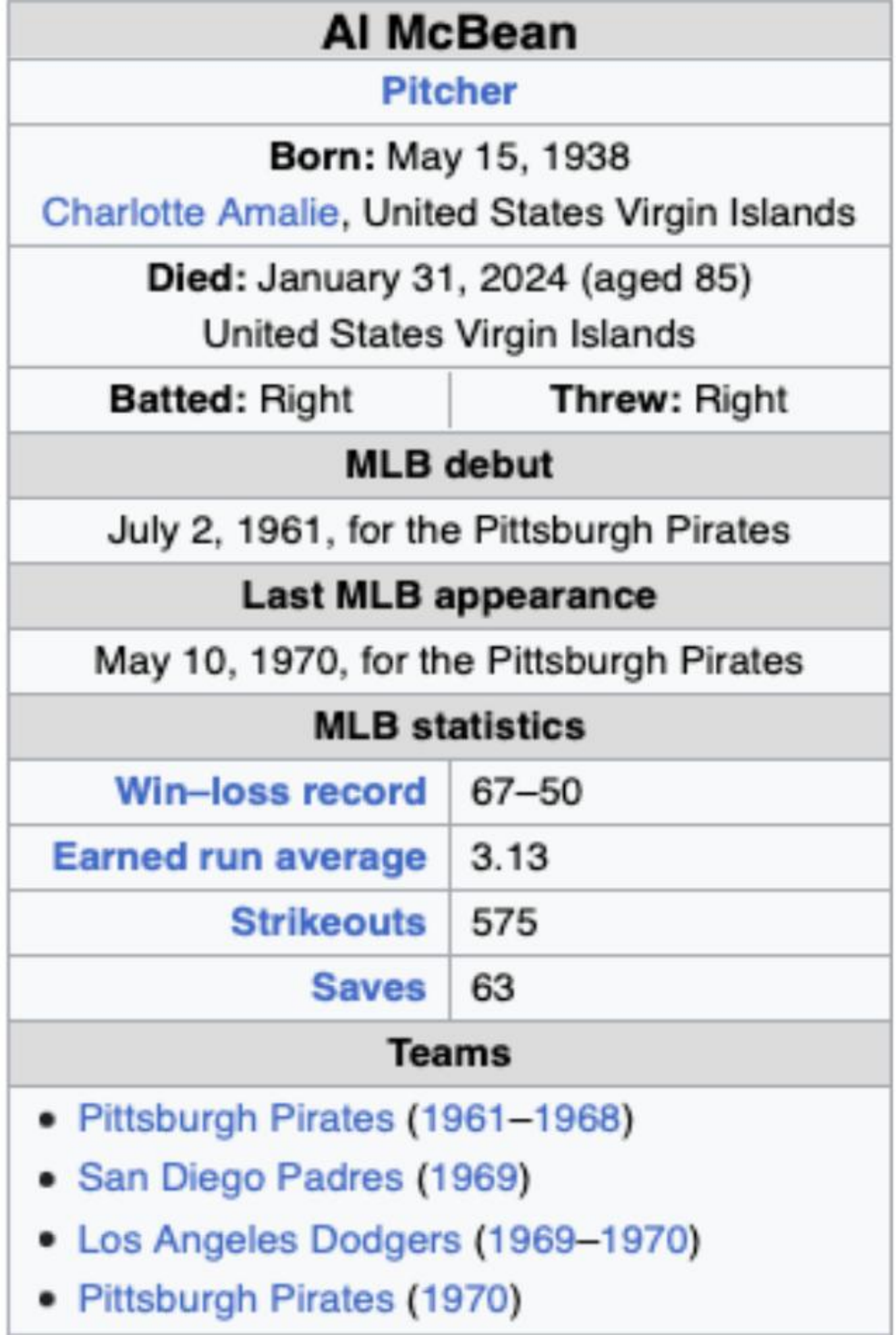}

\vspace{0.25em}
\raggedright
\small{\textbf{Q1 :} How many MLB wins did Al McBean have before turning 20? A: 0 (evidence: Born, MLB statistics)\\
\vspace{0.25em}
\textbf{Q2 :} For how many years did Al McBean play for the Pittsburgh Pirates?  A: 9 (evidence: Teams)}\\
\vspace{0.25em}
\textbf{Q3 :} What was the duration of Al McBean's Major League Baseball career? A: 10 (evidence: MLB debut, Last MLB appearance)
\vspace{-0.5em}
\caption{A semi-structured table of Baseball pitcher Al McBean with follow up question answers.}
  \label{fig:wiki_table}
  \vspace{-2.0em}
\end{figure}

Large Language Models (LLMs) have exhibited remarkable proficiency across various natural language processing tasks. However, recent investigations reveal a notable deficiency in their ability to reason effectively over \emph{tabular} data, particularly when \emph{temporal} relationships are involved \cite{chen-2023-large,sui2024table}. This discrepancy between model performance and human-level understanding underscores the pressing need for innovative approaches to enhance the capabilities of LLMs in this domain.

To explore the root causes behind the limitations in reasoning about structured or semi-structured data, we conducted a thorough analysis of TempTabQA \cite{gupta-etal-2023-temptabqa} dataset, an important and only datasets in this domain. Through meticulous examination, we identified discrepancies in model comprehension and reasoning, leading to the development of an enhanced evaluation set. Our investigation also delved into the mechanisms through which LLMs tackle temporal reasoning tasks using conventional prompting techniques, leasing in the introduction of a novel approach, C.L.E.A.R (Comprehend, Locate, Examine, Analyze, Resolve), designed to considerably enhance temporal reasoning in LLMs.

While prompts have been instrumental in guiding models to better reason about tasks through explicit instructions, they do not \emph{inherently improve} task understanding. In our research, we demonstrate that despite the implementation of our novel C.L.E.A.R approach, models, though improved, still overlook crucial context—such as relevant rows in tables—and suffer from data leakage, relying more on memorization rather than grounding their decisions on the evidence. Additionally, since C.L.E.A.R lacks temporal aspects in its prompt, its effectiveness is limited. This observation necessitate for more robust solutions, where model inherently improve i.e. it's parameters changes.

To address these challenges, we advocate for learning through fine-tuning processes. Fine tuning on temporal data can inherently improve the model’s temporal reasoning ability. We utilized the TRAM \cite{wang2024tram} dataset to engage in an indirectly supervised fine-tuning process by using Auxiliary Out of Domain(OOD) data. This approach significantly improved model \emph{cross-generalization} performance on temporal questions on tabular data. Our findings indicate that combining our novel C.L.E.A.R prompting approach with fine-tuning on the TempTabQA test-set yields optimal results. This combination effectively addresses the limitations in LLMs' temporal reasoning capabilities, paving the way for future advancements in the field. In summary, this paper advances the understanding and enhancement of Large Language Models' (LLMs) abilities to reason about temporal relationships in tabular data. Our key contributions are as follows:
\begin{itemize}
    \item We conducted a thorough analysis of the TempTabQA test set to assess the limitations of current approaches. This analysis also led to an improved test set, enhancing the evaluation of models in temporal tabular reasoning.
    \item We introduce an novel approach, C.L.E.A.R (Comprehend, Locate, Examine, Analyze, Resolve), designed to significantly enhance temporal reasoning capabilities in LLMs for tabular data. Our approach grounds models in evidence, thus reducing memorization.
    \item Through indirect supervision technique a.k.a. auxiliary task training, we inherently improve model performance. Fine-tuning models with the auxiliary TRAM dataset leads to substantial enhancements in handling temporal questions on tabular data.
\end{itemize}

These contributions greatly enhance LLMs' temporal reasoning capabilities, paving the way for future research and applications in temporal reasoning over tabular data. \emph{Our code and enhanced evaluation set will be released for public use.}

\section{Temporal Reasoning on Tables}

Tables organize and record diverse types of information, making them useful for studying an entity’s timeline. They provide a chronological sequence of events, facilitating the analysis of the progression of positions, marital status changes, and awards, thereby serving as reliable sources for temporal reasoning. Entity-centric tables, like Wikipedia Infoboxes, significantly differ from unstructured data and fully structured data (e.g., SQL tables and knowledge graphs).

Recent studies show that Large Language Models (LLMs) struggle with reasoning over tabular data, especially concerning temporal aspects. Various datasets have been used to explore LLMs' understanding of tabular data, including TempTabQA \cite{gupta-etal-2023-temptabqa}, Table2vec \cite{Zhang_2019}, TAPAS \cite{Herzig_2020}, TaBERT \cite{yin2020tabert}, TabStruc \cite{zhang-etal-2020-table}, TABBIE \cite{iida2021tabbie}, TabGCN \cite{pramanick-bhattacharya-2021-joint}, and RCI \cite{glass-etal-2021-capturing}. These datasets provide various contexts and challenges, aiding in the development and evaluation of models for tabular data processing.

Among these datasets, we focus on TempTabQA, a recent and prominent dataset for temporal reasoning over tabular data. TempTabQA features temporal question-answer pairs derived from Wikipedia Infobox tables and is notable for its two distinct eval sets: the "Head" set, with pairs from popular and frequent domains, and the "Tail" set, with pairs from less common and rare domains. The eval sets include approximately 2,900 question-answer pairs, with around 1,900 in the "Head" set and 1,000 in the "Tail" set. This bifurcation enables comprehensive evaluation, assessing models' performance across both frequent and rare domains.

\section{Where do LLMs Fails?}

We used Chain of Thought prompting with GPT-3.5 to analyze the TempTabQA test set and evaluate current models' performance and limitations. Out of 1,038 examples, 339 were incorrect responses. Of these, 159 errors were due to data issues, while 180 were due to model limitations. These errors fall into the following categories:

    \vspace{0.5em}
    \noindent \textbf{1. Tabular Data Issues (75 examples):} These errors were related to hallucinations, incomplete evidence extraction, missing evidence, or incorrect information extraction.

     \vspace{0.5em}
    \noindent \textbf{2. Temporal Calculation Errors (84 examples):} These involved difficulties with calculations related to time, such as determining age, calculating the time between dates in different months, or assessing whether a value fell within a specified range.

     \vspace{0.5em}
    \noindent \textbf{3. Other Errors (31 examples):} This category included errors stemming from common-sense reasoning, arithmetic, and other miscellaneous issues.

\vspace{0.25em}
Our observations indicate that even with Chain of Thought reasoning, models generate incorrect responses consistently. The models not only produce incorrect answers but also exhibit hallucinations, struggling with temporal calculations and common-sense reasoning. This emphasizes the need for enhance models performance in this domain. The 159 data-related issues fall into these categories:
    
    \vspace{0.5em}
    \noindent \textbf{1. Tables Requiring External Knowledge to Answer (75 examples):} These questions could not be answered correctly without additional information not present in the table, indicating a gap in the information provided in the context.
    
    \vspace{0.5em}
    \noindent \textbf{2. Wrong Human Annotation or Multiple Correct Answers (42 examples):} These instances involved incorrect annotations by humans or questions that had multiple valid answers, but the annotations provided only one correct answer.
    
    \vspace{0.5em}
    \noindent \textbf{3. Ambiguous or Incomplete Questions (14 examples):} These questions are either vague or lacked sufficient detail required for correct answer.
    
    \vspace{0.5em}
    \noindent \textbf{4. Other Issues (28 examples):} This category included various problems such as questions relying on images within the HTML table or missing rows in the JSON table etc.

\vspace{0.5em}
Through our analysis, we addressed all these categories of errors and created a new evaluation set designed to better assess model performance. This refined dataset eliminates the noise caused by the aforementioned issues, providing a more accurate benchmark for evaluating model capabilities. In this paper, when evaluating a model’s performance, we primarily use the model’s accuracy on the full test dataset as a basis for comparison, unless otherwise stated. This approach ensures consistency and allows for a clear assessment of improvements made through the new evaluation set.

\section{Methodology}

Improving the temporal reasoning capabilities of Large Language Models is crucial for enhancing their performance on tasks involving time-based data. Current models often struggle with accurately interpreting and reasoning about temporal information. This limitation reduces their effectiveness in real-world applications.

Our analysis revealed several key areas where current models underperform in reasoning over tabular data, especially in temporal contexts. To address these challenges, we explored two approaches. First, we developed a novel method called C.L.E.A.R (Comprehend, Locate, Examine, Analyze, Resolve). This method was inspired by our detailed examination of the TempTabQA test set. C.L.E.A.R instructs models specifically for tabular data reasoning tasks. It employs advanced prompting techniques to guide models more effectively, aiming to reduce errors and enhance accuracy in temporal question answering. Secondly, we fine-tuned models using auxiliary temporal data from unrelated domains to enhance cross-generalization. This approach boosts temporal understanding on complex temporal tasks.

\subsection{C.L.E.A.R Prompting}

In this section, we introduce the C.L.E.A.R (Comprehend, Locate, Examine, Analyze, Resolve) technique, designed to enhance temporal reasoning over semi-structured data. C.L.E.A.R is a structured, step-by-step approach that guides the process of understanding, identifying, examining, analyzing, and resolving questions involving temporal reasoning. It ensures a comprehensive and logical extraction of information from tables (refer Fig. \ref{fig:prompt_example}).

\vspace{0.5em}
\noindent\textbf{(a.) Comprehend:} The first step, Comprehend, involves applying domain knowledge to understand the given question. For example, if the question pertains to calculating time differences, it is essential to recognize that this involves subtracting the earlier year from the later year. This step ensures the correct interpretation of the question and sets the foundation for correct structural solution analysis.

\vspace{0.5em}
\noindent\textbf{(b.) Locate:} The Locate step focuses on identifying and extracting the relevant rows from the table that directly pertain to the question. This involves explaining the rationale behind the selection of these rows to provide transparency and clarity. For instance, if the question asks for events between specific years, only the rows corresponding to those years should be selected. The output of this step includes listing the relevant rows on new lines for clarity and ease of reference.

\vspace{0.5em}
\noindent\textbf{(c.) Examine:} In the Examine phase, the main question is broken down into smaller, more manageable sub-questions. Each sub-question aims at extracting a specific piece of information from the table that is necessary to answer the main question. This decomposition allows for a systematic approach to solving the question, ensuring that no critical piece of information is overlooked.

\vspace{0.5em}
\noindent\textbf{(d.) Analyze:} The Analyze step is multifaceted, involving several sub-steps:

\begin{enumerate}
\setlength\itemsep{0em}
    \item \textit{Mark Evidence for Each Sub-Question: }For each sub-question, identify the specific evidence from the table that will be used to answer it. Explain the relevance of this evidence to the sub-question in detail.
    \item \textit{Determine Dependencies:} This sub-step assesses whether the answer to a sub-question depends on information from previous sub-questions. If dependencies exist, they are stated to maintain logical coherence.
    \item \textit{Reasoning for Each Sub-Question:} A logical explanation is provided for how the evidence leads to the answer for each sub-question. This includes detailed reasoning for transparency and to support the answer's validity.
\end{enumerate}

\noindent\textbf{(e.) Resolve:} The final step, Resolve, involves combining and applying the answers from the sub-questions to formulate the final answer to the main question. This step includes explaining the reasoning process leading to the final answer, which may involve necessary calculations or logical deductions. This synthesis ensures the final answer is well-supported and logically derived from evidence gathered in preceding steps. Refer Figure \ref{fig:clear_working} in Appendix \ref{sec:appendix} for step-by-step process of C.L.E.A.R.

\subsection{Fine Tuning with Auxiliary Data}

Fine-tuning models can significantly boost their capabilities by adjusting parameters through training on task-specific examples. In this section, we demonstrate how models' temporal reasoning can be enhanced inherently, not only through fine-tuning on specific data but also by integrating auxiliary data sources.

Auxiliary data, such as temporal unstructured data, does not directly relate to the main task but contains relevant logical structures and principles. This type of data enhances the model's understanding of underlying logic, improving the overall performance. The TRAM dataset, illustrated in Figure \ref{fig: tram_examples} in  Appendix \ref{sec:appendix}, serves as an auxiliary source to boost model performance. It includes temporal questions that aid the model in grasping temporal relationships across diverse contexts.

The TRAM dataset, though unrelated to tabular data temporal reasoning, enhances model temporal reasoning skills. Exposing the model to diverse temporal questions improves its ability to handle temporal relationships, boosting performance. This approach shows leveraging auxiliary data can effectively address temporal reasoning complexities and enhance model robustness in diverse scenarios.

\section{Experimental Setup}
\label{ref:exepriment_setup}

\textbf{Models:} In this paper, we experimented with several state-of-the-art large language models (LLMs), including GPT-3.5-Turbo, GPT-4, PaLM-2, Mistral-2-7B, LLaMA-2-7B-chat, and Gemini 1.5 Pro Flash. These models represent the forefront in both open-source and closed-model applications, showcasing advancements in natural language understanding and generation capabilities.

\vspace{0.5em}
\noindent\textbf{Prompts:} Prompting models with detailed instructions enhances their understanding of tasks, leading to improved responses. These prompts may include demonstrations for the model's reference. We explore the following prompting techniques:

    \vspace{0.5em}
    \noindent \textbf{- Chain of Thought (CoT)~\cite{wei2023chainofthought}:} CoT prompts guide the model through a series of interconnected thoughts or reasoning steps, encouraging coherent and structured responses. We apply CoT in both zero-shot (Z.S) and few-shot (F.S) settings to evaluate its impact on model performance.

    \vspace{0.5em}
    \noindent \textbf{- Faithful Chain of Thought (F-CoT)~\cite{lyu-etal-2023-faithful}:} F-CoT extends CoT by emphasizing fidelity to the initial prompt throughout response generation to maintain consistency and accuracy in model outputs. We apply F-CoT in both zero-shot and few-shot scenarios to evaluate its effectiveness.

    \vspace{0.5em}
    \noindent \textbf{- Program of Thought (PoT)~\cite{chen2023program}:} PoT provides the model with a predefined sequence of operations, akin to a program, to generate structured and task-specific responses. Like CoT and F-CoT, PoT is evaluated in zero-shot and few-shot contexts to enhance model performance.

\vspace{0.5em}
\noindent\textbf{Auxiliary Data:} In our experiments, we utilize several unstructured temporal reasoning datasets to assess the temporal reasoning capabilities of language models:
    
    \vspace{0.5em}
    \noindent \textbf{- DATE Understanding~\cite{srivastava2023imitation}:} This dataset evaluates a model's ability to comprehend, manipulate, and reason about dates in diverse formats and contexts. Tasks include Date Format Conversion, Date Arithmetic, Date Recognition, Relative Date Interpretation, and Time Reasoning.
    
    \vspace{0.5em}
    \noindent \textbf{- Temporal Sequences~\cite{srivastava2023imitation}:} This dataset tests language models on logical deduction tasks. Models are given a series of events and their durations, and they deduce the possible timing of additional activities.
    
    \vspace{0.5em}
    \noindent \textbf{- TRAM dataset~\cite{wang2024tram}:} This benchmark dataset comprises ten distinct datasets, each focusing on different temporal aspects: order, arithmetic, frequency, and duration. The TRAM dataset aims to evaluate language models' temporal reasoning capabilities comprehensively. Table \ref{fig: tram_examples} shows examples of questions from the TRAM dataset, highlighting the variety and complexity of temporal reasoning tasks included.

\section{Results and Analysis}
In this section, we present the results of enhancing temporal reasoning in Large Language Models (LLMs) for tabular data tasks. We evaluate the effectiveness of two strategies: C.L.E.A.R (Comprehend, Locate, Examine, Analyze, Resolve), a novel method tailored for tabular data reasoning, and the integration of out-of-domain temporal data for model fine-tuning. 

\subsection{C.L.E.A.R Prompting}
\label{sec:results_clear_prompting}
We analyzed C.L.E.A.R prompting against effective techniques like CoT, F-CoT, and PoT in zero-shot and few-shot settings. Our goal was to evaluate their performance across various models and determine the most effective approach.

\begin{table}[!htbp]
\centering
\scriptsize
\setlength{\tabcolsep}{5.85pt}
\begin{tabular}{lcccccc}\toprule
\multirow{2}{*}{\textbf{Prompt}} &\multirow{2}{*}{\textbf{No FT}} &\multicolumn{2}{c}{\textbf{TRAM}} &\multicolumn{2}{c}{\textbf{TempTabQA}} \\\cmidrule{3-6}
& &\textbf{100} &\textbf{1000} &\textbf{100} &\textbf{1000} \\\midrule
& \multicolumn{5}{c}{\textbf{GPT-3.5 turbo}} \\
\textbf{C.L.E.A.R} &\textbf{77.99\%} &\textbf{77.92\%} & 76.12\% &\textbf{80.89\%} &\textbf{82.04\%} \\
\textbf{F.S. CoT} &73.49\% &75.53\% &\textbf{76.19\%} &77.81\% &79.47\% \\
\textbf{F.S. F-CoT} &70.54\% &71.13\% &72.72\% &77.85\% &76.57\% \\
\textbf{F.S. PoT} &70.13\% &70.58\% &69.75\% &79.37\% &77.57\% \\
\textbf{Z.S. CoT} &65.84\% &68.64\% &69.61\% &73.73\% &72.72\% \\
\textbf{Z.S. F-CoT} &64.38\% &67.32\% &65.91\% &71.27\% &71.69\% \\
\textbf{Z.S. PoT} &65.39\% &67.39\% &68.05\% &72.17\% &72.76\% \\
\midrule
& \multicolumn{5}{c}{\textbf{LLAMA 7B}}\\
\textbf{C.L.E.A.R} &59.33\% &61.65\% &59.85\% &63.69\% &63.17\% \\
\textbf{F.S. CoT }&61.51\% &64.87\% &63.66\% &67.22\% &\textbf{68.64\%} \\
\textbf{F.S. F-CoT }&61.96\% &61.54\% &64.24\% &66.77\% &67.32\% \\
\textbf{F.S. PoT} &62.69\% &64.04\% &66.08\% &\textbf{67.43\%} &65.94\% \\
\textbf{Z.S. CoT }&61.65\% &62.48\% &61.79\% &63.07\% &65.21\% \\
\textbf{Z.S. F-CoT} &\textbf{63.93\%} &\textbf{66.98\%} &\textbf{69.40\%} &63.38\% &64.31\% \\
\textbf{Z.S. PoT}&63.34\% &63.69\% &64.76\% &65.07\% &65.42\% \\
\midrule
& \multicolumn{5}{c}{\textbf{PALM 2}} \\
\textbf{C.L.E.A.R} &\textbf{80.06\%} &\textbf{81.97\%} &\textbf{83.28\%} &\textbf{85.01\%} &\textbf{85.01\%} \\
\textbf{F.S. CoT }&77.40\% &81.14\% &82.24\% &80.17\% &81.59\% \\\textbf{F.S. F-CoT} &75.63\% &75.35\% &78.82\% &82.45\% &82.21\% \\
\textbf{F.S. PoT} &76.01\% &75.98\% &76.25\% &84.56\% &80.27\% \\
\textbf{Z.S. CoT} &70.13\% &73.97\% &74.42\% &78.89\% &76.74\% \\
\textbf{Z.S. F-CoT} &69.75\% &74.04\% &74.90\% &75.22\% &75.94\% \\
\textbf{Z.S. PoT} &71.24\% &74.32\% &76.29\% &79.09\% &79.13\% \\
\midrule
& \multicolumn{5}{c}{\textbf{MISTRAL-2 7B}} \\
\textbf{C.L.E.A.R} &\textbf{69.33\%} &\textbf{73.52\%} &\textbf{72.41\%} & 74.80\% &73.87\% \\
\textbf{F.S. CoT} &69.23\% &70.61\% &71.13\% &73.31\% &\textbf{75.63\%} \\
\textbf{F.S. F-CoT} &68.19\% &71.06\% &70.02\% &\textbf{74.84\%} &75.25\% \\
\textbf{F.S. PoT} &67.60\% &70.51\% &69.99\% &74.35\% &73.59\% \\
\textbf{Z.S. CoT} &65.25\% &66.22\% &70.37\% &71.62\% &71.48\% \\
\textbf{Z.S. F-CoT} &66.56\% &71.13\% &69.51\% &67.08\% &71.10\% \\
\textbf{Z.S. PoT} &64.31\% &66.63\% &66.98\% &71.69\% &71.65\% \\
\bottomrule
\end{tabular}
\vspace{-1.0em}
\caption{\small Model's performance on various prompts without fine-tuning, with fine-tuning on auxiliary data (TRAM), and with fine-tuning on the TempTabQA dataset using 100 and 1000 examples. The best result is highlighted in bold.}\label{tab: fine_tuning}
\vspace{-1.75em}
\end{table}

As shown in Table \ref{tab: fine_tuning}, C.L.E.A.R consistently outperforms other techniques across various models without fine-tuning, except for LLaMA-2. For GPT-3.5 Turbo, our method achieves a 4.5\% performance boost compared to the next best technique. For PaLM-2, it leads to a 2.7\% improvement.

These findings highlight C.L.E.A.R prompting's effectiveness in enhancing model performance, particularly in scenarios where fine-tuning is not feasible. Significant performance gains observed with GPT-3.5 Turbo and PaLM-2 demonstrate its potential to improve temporal reasoning and overall task comprehension in large language models.

\subsection{Efficacy of C.L.E.A.R}

The results in Section \ref{sec:results_clear_prompting} demonstrate that our method is more effective than other prompting techniques. However, is C.L.E.A.R \emph{trustworthy?} Does it really enhance the model's \emph{evidence-based reasoning capabilities?}~\cite{table-probing}. Our experiments compare C.L.E.A.R with Zero Shot and Few Shot Chain of Thought (CoT) approaches across tasks highlighting model deficiencies. Below are each of task descriptions:
\begin{enumerate}
\setlength\itemsep{0em}
\vspace{-0.5em}
    \item \textbf{Original Table:} The model uses the original table to answer the question, testing its ability to effectively utilize the provided data.
    \item \textbf{Without Table:} The model is asked to answer the question without access to the table, testing whether the model has memorized the answers or can deduce them independently. Here, an ideal model's performance should suffer as it doesn't use pre-trained knowledge.
    \item \textbf{Altered Entity Name:} The table and question are provided to the model with an altered entity name. This checks the model's reasoning ability without relying on memorized data.
    \item \textbf{Missing Relevant Rows:} The model is given the original table with relevant rows deleted. This evaluates whether the model can use external knowledge to answer questions. Responses are evaluated against gold answers.
    \item \textbf{Information Absence Detection:} The model receives the original table with relevant rows deleted. This task evaluates the model's accuracy in identifying missing information.
\end{enumerate}

\noindent For the last two tasks i.e. Missing Relevant Rows and Information Absence Detection, we perform evaluations in two settings:
\begin{itemize}
\vspace{-0.5em}
\setlength\itemsep{0em}
    \item \textbf{Original Prompt:} The model is tested with the original prompt to see if it can detect missing information or use external knowledge.
    \item \textbf{Updated Prompt:} The model is explicitly instructed that the information may or may not be present, to see if explicit instructions improve performance.
\end{itemize}

We aim to assess if C.L.E.A.R prompting improves evidence-based reasoning in models beyond accuracy. This analysis evaluates its effectiveness in addressing specific reasoning challenges and enhancing model reliability across diverse contexts.

\paragraph{Results and Analysis.} We tested zero-shot Chain of Thought (CoT), few-shot CoT, and C.L.E.A.R prompting techniques across various tasks. In Table \ref{tab: clear_tasks}, C.L.E.A.R consistently outperforms other methods on most tasks, except for LLaMA-2 and Mistral-2 models, which show varied results. GPT 4O achieves the highest accuracy at 85.08\%, marking a 2.39\% improvement over few-shot CoT.

\vspace{0.25em}
\textbf{(a) Without Table.} In this task, when evaluated without table access, C.L.E.A.R's performance declines across all models using original labels. This indicates that it mitigates memorization issues seen in other techniques by emphasizing contextual reliance. Specifically, we observe a decrease of  \url{~}4\% for GPT-4O and GPT-3.5 Turbo, and an 11.04\% decrease for Llama.

\vspace{0.25em}
\textbf{(b) Altered Table Entity.} C.L.E.A.R performs best overall in this task, except for LLaMA-2, PaLM, and Mistral-2. GPT 4O provides best accuracy of 82.62\% which is a bosst of 6.26\% in comparision of few shot CoT. LLaMA-2 7B and Mistral-2 7B show weaker performance mainly due to their smaller size, which generally leads to less effective adherence to instructions compared to larger models. For models GPT 3.5 turbo and Gemini 1.5 pro flash, we observe a gain of 4.15\% and 8.51\% respectively.

\begin{table}[!htp]\centering
\scriptsize
\setlength{\tabcolsep}{2.7pt}
\begin{tabular}{llrrrr}\toprule
\multirow{2}{*}{\textbf{Models}} &\multirow{2}{*}{\textbf{Task (expected)}} &\multicolumn{3}{c}{\textbf{Prompts}} \\\cmidrule{3-5}
& &\textbf{Z.S CoT} &\textbf{F.S CoT} &\textbf{C.L.E.A.R} \\\midrule
\multirow{3}{*}{\textbf{GPT 4O}} &\textbf{Original Table ($\uparrow$)} &75.80\% &82.69\% &\textbf{85.08\%} \\
&\textbf{Without Table ($\downarrow$)} &75.91\% &73.31\% &\textbf{69.44\%} \\
&\textbf{Altered Name Entity ($\uparrow$)} &74.94\% &76.36\% &\textbf{82.62\%} \\
\textbf{} &\textbf{} & & & \\
\multirow{3}{*}{\textbf{GPT 3.5}} &\textbf{Original Table ($\uparrow$)} &65.56\% &73.35\% &\textbf{78.05\%} \\
&\textbf{Without Table ($\downarrow$)} &62.27\% &62.10\% &\textbf{58.22\%} \\
&\textbf{Altered Name Entity($\uparrow$)} &61.09\% &67.95\% &\textbf{72.10\%} \\
\textbf{} &\textbf{} & & & \\
\multirow{3}{*}{\textbf{Llama}} &\textbf{Original Table ($\uparrow$)} &62.31\% &\textbf{62.41\%} &59.12\% \\
&\textbf{Without Table ($\downarrow$)} &54.69\% &42.85\% &\textbf{31.81\%} \\
&\textbf{Altered Name Entity} &\textbf{66.87\%} &60.78\% &58.15\% \\
\textbf{} &\textbf{} & & & \\
\multirow{3}{*}{\textbf{Palm 2}} &\textbf{Original Table ($\uparrow$)} &69.75\% &77.29\% &\textbf{79.58\%} \\
&\textbf{Without Table ($\downarrow$)} &66.63\% &63.24\% &\textbf{60.26\%} \\
&\textbf{Altered Name Entity ($\uparrow$)} &\textbf{82.10\%} &71.34\% &74.97\% \\
\textbf{} &\textbf{} & & & \\
\multirow{3}{*}{\textbf{MISTRAL-2}} &\textbf{Original Table ($\uparrow$)} &64.87\% &\textbf{69.37\%} &69.30\% \\
&\textbf{Without Table ($\downarrow$)} &56.87\% &52.58\% &\textbf{49.36\%} \\
&\textbf{Altered Name Entity ($\uparrow$)} &\textbf{74.87\%} &63.55\% &64.83\% \\
\textbf{} &\textbf{} & & & \\
\multirow{3}{*}{\makecell{\textbf{GEMINI 1.5} \\ \textbf{}}} &\textbf{Original Table ($\uparrow$)} &73.17\% &80.34\% &\textbf{83.28\%} \\
&\textbf{Without Table ($\downarrow$)} &70.85\% &69.19\% &\textbf{68.64\%} \\
&\textbf{Altered Name Entity ($\uparrow$)} &79.02\% &71.24\% &\textbf{79.75\%} \\
\bottomrule
\end{tabular}
\vspace{-0.75em}
\caption{\small This table provides a comparison among C.L.E.A.R, zero-shot CoT, and few-shot CoT prompts across Original Table, No Table, and Altered Named Entity tasks in evidence-based reasoning. We evaluate performance using GPT-4o, GPT-3.5 Turbo, LLAMA-7B-Chat, PaLM-2, MISTRAL-7B, and Gemini 1.5 Pro Fash models in our experiments.}\label{tab: clear_tasks}
\vspace{-1.5em}
\end{table}

\textbf{(c) Missing Relevant Row.} We observe in Table \ref{tab: clear_tasks2}, C.L.E.A.R outperforms others across all models with the original prompt. It shows the largest performance decline compared to zero-shot and few-shot CoT. This trend continues with the modified prompt, where our method remains superior, except for GPT-4, where zero-shot performs slightly better. Prompting creates an additional drop of \url{~}1-5\% in all models. Explicitly informing the model about the availability of information promotes context-driven responses over reliance on memorized data.

\textbf{(d) Information Absence Detection} In Table \ref{tab: clear_tasks2}, our method outperforms zero-shot and few-shot CoT with the original prompt across all models. Llama achieves the highest accuracy at 40.05\%, followed by GPT-3.5 at 31.71\%. With the updated prompt, our technique consistently performs best for all models. Gemini 1.5 Pro Flash achieves the highest accuracy at 91.45\%, with GPT-4O close behind at 90.65

\begin{table}[!tb]
\centering
\scriptsize
\setlength{\tabcolsep}{2.1pt}
\begin{tabular}{l|ccc|ccc}\toprule
{\textbf{Models}} &\multicolumn{3}{c|}{\textbf{Original Prompt}} &\multicolumn{3}{c}{\textbf{Updated Prompt}} 
\\\cmidrule{1-7}
& \multicolumn{2}{c}{\textbf{CoT}} & & \multicolumn{2}{c}{\textbf{CoT}} & 
\\\cmidrule{2-3} \cmidrule{5-6}
&\textbf{Z.S } &\textbf{F.S} & \textbf{C.L.E.A.R} &\textbf{Z.S} &\textbf{F.S} &  \textbf{C.L.E.A.R} \\\cmidrule{1-7}
& \multicolumn{6}{c}{\textbf{Missing Relevant Rows (Lower ($\downarrow$) is better)}} \\\midrule
\textbf{GPT 4O} &69.71\% &70.89\% &\textbf{62.82\%} &\textbf{6.89\%} &8.34\% &7.30\% \\
\textbf{GPT 3.5} &60.06\% &56.59\% &\textbf{49.46\%} &11.60\% &13.05\% &\textbf{11.39\%} \\
\textbf{LLAMA 7B} &64.42\% &37.69\% &\textbf{24.92\%} &26.10\% &16.13\% &\textbf{11.18\%} \\
\textbf{PALM 2} &78.40\% &61.68\% &\textbf{55.90\%} &10.90\% &8.27\% &\textbf{7.58\%} \\
\textbf{MISTRAL-2} &69.92\% &50.85\% &\textbf{39.84\%} &16.44\% &13.08\% &\textbf{10.76\%} \\
\textbf{GEMINI 1.5} &76.25\% &68.05\% &\textbf{57.29\%} &9.07\% &8.45\% &\textbf{7.34\%} \\\midrule
& \multicolumn{6}{c}{\textbf{Information Absence Detection (Higher ($\uparrow$) is better}} \\\midrule
\textbf{GPT 4O} &19.45\% &22.71\% &\textbf{25.16\%} &89.89\% &88.85\% &\textbf{90.65\%} \\
\textbf{GPT 3.5} &27.83\% &29.73\% &\textbf{31.71\%} &79.75\% &79.47\% &\textbf{82.07\%} \\
\textbf{LLAMA 7B} &25.37\% &34.96\% &\textbf{40.05\%} &61.72\% &61.96\% &\textbf{62.34\%} \\
\textbf{PALM 2} &19.21\% &23.68\% &\textbf{29.11\%} &86.43\% &86.78\% &\textbf{88.44\%} \\
\textbf{MISTRAL-2} &21.05\% &29.42\% &\textbf{32.71\%} &75.42\% &74.49\% &\textbf{77.09\%} \\
\textbf{GEMINI 1.5} &15.82\% &22.33\% &\textbf{25.23\%} &89.48\% &87.61\% &\textbf{91.45\%} \\
\bottomrule
\end{tabular}
\vspace{-0.5em}
\caption{\small This table compares C.L.E.A.R, zero-shot CoT, and few-shot CoT prompts for tasks involving Missing Relevant Rows and Information Absence Detection in evidence-based reasoning, considering both original and changed prompts. Experimental evaluations are conducted using GPT-4o, GPT-3.5 Turbo, LLAMA-7B-Chat, PaLM-2, MISTRAL-7B, and Gemini 1.5 Pro Fash models.}\label{tab: clear_tasks2}
\vspace{-2.0em}
\end{table}

\subsection{Auxiliary Data Fine Tuning}
C.L.E.A.R enhances the model's capacity to process and reason with context. It facilitates effective instruction for models to handle temporal questions and improve evidence-based reasoning. However, prompting alone does not adjust their inherent parameters, so models do not intrinsically improve. Innate enhancement in models requires fine-tuning, so that their parameters better reflect improved temporal reasoning capabilities. We suggest fine-tuning the model on simple auxiliary data to enhance its ability to reason across diverse data formats and improve overall reasoning capabilities.

\paragraph{Why TRAM dataset?} We evaluated the impact of fine-tuning GPT-3.5 Turbo using the auxiliary datasets discussed in Section \ref{ref:exepriment_setup}, along with the TempTabQA dataset. For this evaluation, we fine-tuned with 100 examples from each dataset. 

\begin{table}[!htp]
\vspace{-0.5em}
\centering
\scriptsize
\begin{tabular}{lc}\toprule
\textbf{Train Dataset} & \textbf{Exact Match} \\\midrule
No fine-tuning &73.49\% \\
DATE &74.28\% \\
Temporal Sequences &74.42\% \\
TRAM &75.53\% \\
TempTabQA &77.81\% \\
\bottomrule
\end{tabular}
\vspace{-0.75em}
\caption{\small This table showcases the performance of GPT-3.5 Turbo following fine-tuning on auxiliary datasets within a zero-shot chain-of-thought (CoT) setting."}\label{tab: auxi_datasets}
\vspace{-0.75em}
\end{table}

\textit{Analysis.} Table \ref{tab: auxi_datasets} shows that the TRAM dataset offers the largest performance improvement among auxiliary datasets. While DATE and Temporal Sequences datasets provide gains of less than 1.01\%, TRAM achieves a significant 2.04\% boost compared to the base model without fine-tuning. These findings emphasize TRAM's unique advantages. DATE focuses solely on date-based reasoning, and Temporal Sequences on temporal reasoning across events. In contrast, TRAM includes 10 sub-datasets covering diverse temporal reasoning tasks. It includes tasks such as \emph{Ordering, Frequency, Duration, Typical time, Ambiguity Resolution, Arithmetic, Relations, Temporal NLI, Causality and Storytelling}. Each task have multiple question type including \emph{Commonsense, Facts, Reading Comprehension, Application, Computation, Direct \& Multi-step Comparison, Interpretation, Calender shift, Long-term, Mid-term \& Short-term shift, Date Computation, 12 \& 24 hour adjustment, Week Identification etc}. (For more information on TRAM, refer to Table \ref{tab: tram_overview} in Appendix \ref{sec:appendix}). This diversity enhances performance by exposing models to varied temporal scenarios, improving overall comprehension in temporal domain.

The TRAM dataset's significant gains highlight the importance of diverse auxiliary data for fine-tuning. By covering a wide range of temporal reasoning tasks, TRAM helps models grasp nuanced temporal relationships, enhancing performance across various challenges. This finding suggests that future efforts in model fine-tuning should consider leveraging similarly diverse datasets to maximize performance improvements.

\paragraph{Fine-tuning on TRAM.} We analyzed fine-tuning models on subsets of the TRAM dataset and TempTabQA, each with 100 and 1000 examples (evenly sampled across tasks), as shown in Table \ref{tab: fine_tuning}.

\textit{Analysis.} Our findings reveal that, on the TRAM dataset with fine-tuning on 100 examples, C.L.E.A.R outperforms all models except for LLAMA-2 7B. This trend is consistent with 1000 examples, where it slightly trails behind the few-shot CoT for GPT-3.5 turbo by a negligible margin of 0.07\%.When fine-tuning models on TempTabQA with 100 examples and 1000 examples, our method demonstrates superior performance for GPT 3.5 turbo and PaLM-2, but underperforms for LLAMA-2 7B and Mistral-2 7B.

Our findings show that fine-tuning models with auxiliary data significantly improves performance, close to that of task-specific fine-tuning. This method enhances models' ability to reason about temporal information and can be applied across various tasks. Additionally, auxiliary data typically provides richer resources compared to task-specific datasets. This scalability enables fine-tuning on larger datasets beyond traditional methods. \footnote{Due to computational limits, especially with GPT and Gemini models, we fine-tuned LLMs on just 1000 examples.}

\section{Comparison with Related Work}

\paragraph{Tabular Reasoning} Recent research has extensively explored large language models' applications with semi-structured tabular data \cite{Chen2020TabFact:, gupta-etal-2020-infotabs,Zhang:2019:ADC}, covering areas such as question answering and semantic parsing \cite{Zhang:2020:SET, Zhang:2020:WTE, pasupat2015compositional,krishnamurthy2017neural,Abbas2016WikiQAA,sun2016table,chen2020hybridqa,lin2020bridging,zayats2021representations, oguz2020unified,chen2020open,iyyer-etal-2017-search}, as well as table-to-text generation \cite{parikh2020totto,li-etal-2021-twt-table, radev2020dart,yoran2021turning,chen2020logical}. Various datasets and models have emerged to handle semi-structured data, including Table2vec \cite{Zhang_2019}, TAPAS \cite{Herzig_2020}, TaBERT \cite{yin2020tabert}, TabStruc \cite{zhang-etal-2020-table}, TABBIE \cite{iida2021tabbie}, TabGCN \cite{pramanick-bhattacharya-2021-joint}, and RCI \cite{glass-etal-2021-capturing}, all aimed at improving understanding and representation of tabular data. Research has also focused on fine-tuning models for enhanced inference on tabular data, as demonstrated by studies \cite{yu-etal-2018-spider,eisenschlos-etal-2020-understanding,neeraja-etal-2021-incorporating}, showcasing targeted techniques to boost model performance.

Recently, \cite{srivastava2024evaluating} introduced EEDP, a novel approach for financial document Question Answering. EEDP retrieves finance domain information, extracts relevant table rows, and decomposes mathematical tasks into atomic operations. Our method similarly involves understanding table data and extracting relevant rows, but we differentiate by systematically addressing sub-problems based on extracted evidence. This targeted approach makes our method well-suited for temporal tabular reasoning.

\paragraph{Temporal Reasoning} In temporal question answering, recent datasets like TIME-SENSITIVEQA \citep{chen2021a} and TORQUE \citep{ning-etal-2020-torque} focus on entity-specific reading comprehension with time-sensitive questions. Other datasets like TEMPQA-WD \citep{neelam2022benchmark}, CRONQUESTIONS \citep{saxena-etal-2021-question}, and TEMPQUESTIONS \citep{10.1145/3184558.3191536} explore temporal links in knowledge graph embeddings. There are also open-domain \citep{zhang-choi-2021-situatedqa} and cloze-form \citep{dhingra-etal-2022-time} question-answering tasks, and event-centric datasets \citep{ning-etal-2018-cogcomptime,wen-etal-2021-event,chen2021event}. Models like CRONKBQA \citep{saxena-etal-2021-question}, TEQUILA \citep{jia:18b}, and others \citep{jia2021complex,shang2021open,mavromatis2021tempoqr,kannen2022targeted} have been tailored for knowledge-based question answering with temporal considerations, often integrating temporal aspects during pre-training \citep{dhingra-etal-2022-time, iv-etal-2022-fruit}. Additionally, temporal reasoning across structured and unstructured data has seen advancements with works such as TempTabQA \cite{gupta-etal-2023-temptabqa}, TRAM \cite{wang2024tram}, Temporal Sequencescite \cite{srivastava2023imitation}, and DATE understanding \cite{srivastava2023imitation}, all addressing the challenge of handling temporal information effectively.

Our work focuses on enhancing temporal reasoning in tabular data. We introduce a novel prompting method to improve language models' temporal reasoning abilities with tabular data. Additionally, we demonstrate the effectiveness of augmenting models' temporal understanding using auxiliary data, a novel approach in the field. This combined approach significantly enhances temporal inference, paving the way for future advancements.

\section{Conclusion and Future Work}

Our results demonstrate that C.L.E.A.R (Comprehend, Locate, Examine, Analyze, Resolve) prompting significantly enhances model performance, particularly in understanding tabular data and tasks requiring temporal reasoning. It promotes grounding LLMs in evidence context rather than relying solely on pre-trained knowledge. Moreover, fine-tuning LLMs with auxiliary temporal data has proven highly effective in enhancing temporal understanding. This study validates that integrating auxiliary data strengthens models' temporal reasoning capabilities, thereby enhancing the robustness of large language models. For Future, we propose \textbf{(a.) Generation of Synthetic Data:} Creating synthetic training data from temporal aspects of tabular data enhances model performance through diverse temporal exposure. \textbf{(b.) Neuro-symbolic Learning:} Integrating neural networks with symbolic reasoning improves models' understanding of temporal information for more effective solutions. \textbf{(c.) Expanding C.L.E.A.R Prompting Applications:} Applying C.L.E.A.R prompting to diverse tasks and domains validates its versatility in natural language processing, enhancing reasoning with temporal information. \textbf{(d.) Integration with Existing Models:} Seamlessly integrating C.L.E.A.R prompting and auxiliary data into existing models maximizes benefits without architectural changes.

\section*{Limitations}
The experiments in this paper have been conducted exclusively on the English language. This study can be extended to a multilingual setting to evaluate the approach's effectiveness across different languages. Additionally, the temporal datasets used in our study are limited to simple, entity-centric tables. Since structured data can exist in more complex forms, such as hierarchical tables, further research is necessary to assess the impact of our methods on these more complex structures.

Moreover, our computational limitations restricted us to fine-tuning models on only 1000 samples of auxiliary data. To fully understand the potential improvements from fine-tuning on auxiliary data, it is essential to explore the effects of fine-tuning on larger datasets. Future work should focus on overcoming these limitations to provide a comprehensive evaluation of our approach.

\section*{Ethics Statement}

We confirm that our work adheres to the highest ethical standards in research and publication. We will publicly release our code and enhanced evaluation set to enable the research community to validate and build upon our findings. We are committed to the responsible and fair use of computational linguistics methodologies. The claims in our paper accurately reflect the experimental results. While using black-box large language models introduces some stochasticity, we mitigate this by maintaining a fixed temperature. We utilize an AI assistive tools for writing while ensuring absence of bias. We provide comprehensive details on annotations, dataset splits, models used, and prompting methods tried, ensuring the reproducibility of our work.

\section{Acknowledgement}

Research was sponsored by the Army Research Office and was accomplished under Grant Number W911NF-20-1-0080. The views and conclusions contained in this document are those of the authors and should not be interpreted as representing the official policies, either expressed or implied, of the Army Research Office or the U.S. Government. The U.S. Government is authorized to reproduce and distribute reprints for Government purposes notwithstanding any copyright notation herein. This work was partially funded by ONR Contract N00014-23-1-2365. 

\bibliography{anthology,custom}

\begin{thebibliography}{56}
\expandafter\ifx\csname natexlab\endcsname\relax\def\natexlab#1{#1}\fi

\bibitem[{Abbas et~al.(2016)Abbas, Malik, Rashid, and Zafar}]{Abbas2016WikiQAA}
Faheem Abbas, M.~K. Malik, M.~Rashid, and Rizwan Zafar. 2016.
\newblock Wikiqa — a question answering system on wikipedia using freebase, dbpedia and infobox.
\newblock \emph{2016 Sixth International Conference on Innovative Computing Technology (INTECH)}, pages 185--193.

\bibitem[{Chen et~al.(2021{\natexlab{a}})Chen, Zhang, Ning, Li, Ji, McKeown, and Roth}]{chen2021event}
Muhao Chen, Hongming Zhang, Qiang Ning, Manling Li, Heng Ji, Kathleen McKeown, and Dan Roth. 2021{\natexlab{a}}.
\newblock Event-centric natural language understanding.
\newblock In \emph{Proceedings of the 59th Annual Meeting of the Association for Computational Linguistics}.

\bibitem[{Chen(2023)}]{chen-2023-large}
Wenhu Chen. 2023.
\newblock \href {https://doi.org/10.18653/v1/2023.findings-eacl.83} {Large language models are few(1)-shot table reasoners}.
\newblock In \emph{Findings of the Association for Computational Linguistics: EACL 2023}, pages 1120--1130, Dubrovnik, Croatia. Association for Computational Linguistics.

\bibitem[{Chen et~al.(2021{\natexlab{b}})Chen, Chang, Schlinger, Wang, and Cohen}]{chen2020open}
Wenhu Chen, Ming-Wei Chang, Eva Schlinger, William~Yang Wang, and William~W. Cohen. 2021{\natexlab{b}}.
\newblock \href {https://openreview.net/forum?id=MmCRswl1UYl} {Open question answering over tables and text}.
\newblock In \emph{International Conference on Learning Representations}.

\bibitem[{Chen et~al.(2020{\natexlab{a}})Chen, Chen, Su, Chen, and Wang}]{chen2020logical}
Wenhu Chen, Jianshu Chen, Yu~Su, Zhiyu Chen, and William~Yang Wang. 2020{\natexlab{a}}.
\newblock \href {https://doi.org/10.18653/v1/2020.acl-main.708} {Logical natural language generation from open-domain tables}.
\newblock In \emph{Proceedings of the 58th Annual Meeting of the Association for Computational Linguistics}, pages 7929--7942, Online. Association for Computational Linguistics.

\bibitem[{Chen et~al.(2023)Chen, Ma, Wang, and Cohen}]{chen2023program}
Wenhu Chen, Xueguang Ma, Xinyi Wang, and William~W. Cohen. 2023.
\newblock \href {http://arxiv.org/abs/2211.12588} {Program of thoughts prompting: Disentangling computation from reasoning for numerical reasoning tasks}.

\bibitem[{Chen et~al.(2020{\natexlab{b}})Chen, Wang, Chen, Zhang, Wang, Li, Zhou, and Wang}]{Chen2020TabFact:}
Wenhu Chen, Hongmin Wang, Jianshu Chen, Yunkai Zhang, Hong Wang, Shiyang Li, Xiyou Zhou, and William~Yang Wang. 2020{\natexlab{b}}.
\newblock \href {https://openreview.net/forum?id=rkeJRhNYDH} {Tabfact: A large-scale dataset for table-based fact verification}.
\newblock In \emph{International Conference on Learning Representations}.

\bibitem[{Chen et~al.(2021{\natexlab{c}})Chen, Wang, and Wang}]{chen2021a}
Wenhu Chen, Xinyi Wang, and William~Yang Wang. 2021{\natexlab{c}}.
\newblock \href {https://openreview.net/forum?id=9-LSfSU74n-} {A dataset for answering time-sensitive questions}.
\newblock In \emph{Thirty-fifth Conference on Neural Information Processing Systems Datasets and Benchmarks Track (Round 2)}.

\bibitem[{Chen et~al.(2020{\natexlab{c}})Chen, Zha, Chen, Xiong, Wang, and Wang}]{chen2020hybridqa}
Wenhu Chen, Hanwen Zha, Zhiyu Chen, Wenhan Xiong, Hong Wang, and William~Yang Wang. 2020{\natexlab{c}}.
\newblock \href {https://doi.org/10.18653/v1/2020.findings-emnlp.91} {{H}ybrid{QA}: A dataset of multi-hop question answering over tabular and textual data}.
\newblock In \emph{Findings of the Association for Computational Linguistics: EMNLP 2020}, pages 1026--1036, Online. Association for Computational Linguistics.

\bibitem[{Dhingra et~al.(2022)Dhingra, Cole, Eisenschlos, Gillick, Eisenstein, and Cohen}]{dhingra-etal-2022-time}
Bhuwan Dhingra, Jeremy~R. Cole, Julian~Martin Eisenschlos, Daniel Gillick, Jacob Eisenstein, and William~W. Cohen. 2022.
\newblock \href {https://doi.org/10.1162/tacl_a_00459} {Time-aware language models as temporal knowledge bases}.
\newblock \emph{Transactions of the Association for Computational Linguistics}, 10:257--273.

\bibitem[{Eisenschlos et~al.(2020)Eisenschlos, Krichene, and M{\"u}ller}]{eisenschlos-etal-2020-understanding}
Julian Eisenschlos, Syrine Krichene, and Thomas M{\"u}ller. 2020.
\newblock \href {https://doi.org/10.18653/v1/2020.findings-emnlp.27} {Understanding tables with intermediate pre-training}.
\newblock In \emph{Findings of the Association for Computational Linguistics: EMNLP 2020}, pages 281--296, Online. Association for Computational Linguistics.

\bibitem[{Glass et~al.(2021)Glass, Canim, Gliozzo, Chemmengath, Kumar, Chakravarti, Sil, Pan, Bharadwaj, and Fauceglia}]{glass-etal-2021-capturing}
Michael Glass, Mustafa Canim, Alfio Gliozzo, Saneem Chemmengath, Vishwajeet Kumar, Rishav Chakravarti, Avi Sil, Feifei Pan, Samarth Bharadwaj, and Nicolas~Rodolfo Fauceglia. 2021.
\newblock \href {https://doi.org/10.18653/v1/2021.naacl-main.96} {Capturing row and column semantics in transformer based question answering over tables}.
\newblock In \emph{Proceedings of the 2021 Conference of the North American Chapter of the Association for Computational Linguistics: Human Language Technologies}, pages 1212--1224, Online. Association for Computational Linguistics.

\bibitem[{Gupta et~al.(2021)Gupta, Bhat, Ghosal, Srivastava, Singh, and Srikumar}]{table-probing}
Vivek Gupta, Riyaz~A. Bhat, Atreya Ghosal, Manish Srivastava, Maneesh Singh, and Vivek Srikumar. 2021.
\newblock \href {http://arxiv.org/abs/2108.00578} {Is my model using the right evidence? systematic probes for examining evidence-based tabular reasoning}.
\newblock \emph{CoRR}, abs/2108.00578.

\bibitem[{Gupta et~al.(2023)Gupta, Kandoi, Vora, Zhang, He, Reinanda, and Srikumar}]{gupta-etal-2023-temptabqa}
Vivek Gupta, Pranshu Kandoi, Mahek Vora, Shuo Zhang, Yujie He, Ridho Reinanda, and Vivek Srikumar. 2023.
\newblock \href {https://doi.org/10.18653/v1/2023.emnlp-main.149} {{T}emp{T}ab{QA}: Temporal question answering for semi-structured tables}.
\newblock In \emph{Proceedings of the 2023 Conference on Empirical Methods in Natural Language Processing}, pages 2431--2453, Singapore. Association for Computational Linguistics.

\bibitem[{Gupta et~al.(2020)Gupta, Mehta, Nokhiz, and Srikumar}]{gupta-etal-2020-infotabs}
Vivek Gupta, Maitrey Mehta, Pegah Nokhiz, and Vivek Srikumar. 2020.
\newblock \href {https://doi.org/10.18653/v1/2020.acl-main.210} {{INFOTABS}: Inference on tables as semi-structured data}.
\newblock In \emph{Proceedings of the 58th Annual Meeting of the Association for Computational Linguistics}, pages 2309--2324, Online. Association for Computational Linguistics.

\bibitem[{Herzig et~al.(2020)Herzig, Nowak, Müller, Piccinno, and Eisenschlos}]{Herzig_2020}
Jonathan Herzig, Pawel~Krzysztof Nowak, Thomas Müller, Francesco Piccinno, and Julian Eisenschlos. 2020.
\newblock \href {https://doi.org/10.18653/v1/2020.acl-main.398} {Tapas: Weakly supervised table parsing via pre-training}.
\newblock In \emph{Proceedings of the 58th Annual Meeting of the Association for Computational Linguistics}. Association for Computational Linguistics.

\bibitem[{Iida et~al.(2021)Iida, Thai, Manjunatha, and Iyyer}]{iida2021tabbie}
Hiroshi Iida, Dung Thai, Varun Manjunatha, and Mohit Iyyer. 2021.
\newblock \href {http://arxiv.org/abs/2105.02584} {Tabbie: Pretrained representations of tabular data}.

\bibitem[{Iv et~al.(2022)Iv, Passos, Singh, and Chang}]{iv-etal-2022-fruit}
Robert Iv, Alexandre Passos, Sameer Singh, and Ming-Wei Chang. 2022.
\newblock \href {https://doi.org/10.18653/v1/2022.naacl-main.269} {{FRUIT}: Faithfully reflecting updated information in text}.
\newblock In \emph{Proceedings of the 2022 Conference of the North American Chapter of the Association for Computational Linguistics: Human Language Technologies}, pages 3670--3686, Seattle, United States. Association for Computational Linguistics.

\bibitem[{Iyyer et~al.(2017)Iyyer, Yih, and Chang}]{iyyer-etal-2017-search}
Mohit Iyyer, Wen-tau Yih, and Ming-Wei Chang. 2017.
\newblock \href {https://doi.org/10.18653/v1/P17-1167} {Search-based neural structured learning for sequential question answering}.
\newblock In \emph{Proceedings of the 55th Annual Meeting of the Association for Computational Linguistics (Volume 1: Long Papers)}, pages 1821--1831, Vancouver, Canada. Association for Computational Linguistics.

\bibitem[{Jia et~al.(2018{\natexlab{a}})Jia, Abujabal, Saha~Roy, Str\"{o}tgen, and Weikum}]{10.1145/3184558.3191536}
Zhen Jia, Abdalghani Abujabal, Rishiraj Saha~Roy, Jannik Str\"{o}tgen, and Gerhard Weikum. 2018{\natexlab{a}}.
\newblock \href {https://doi.org/10.1145/3184558.3191536} {Tempquestions: A benchmark for temporal question answering}.
\newblock In \emph{Companion Proceedings of the The Web Conference 2018}, WWW '18, page 1057–1062, Republic and Canton of Geneva, CHE. International World Wide Web Conferences Steering Committee.

\bibitem[{Jia et~al.(2018{\natexlab{b}})Jia, Abujabal, Saha~Roy, Str\"{o}tgen, and Weikum}]{jia:18b}
Zhen Jia, Abdalghani Abujabal, Rishiraj Saha~Roy, Jannik Str\"{o}tgen, and Gerhard Weikum. 2018{\natexlab{b}}.
\newblock \href {https://doi.org/10.1145/3269206.3269247} {{TEQUILA: Temporal Question Answering over Knowledge Bases}}.
\newblock In \emph{Proceedings of the 27th ACM International Conference on Information and Knowledge Management}, CIKM '18, pages 1807--1810, New York, NY, USA. ACM.

\bibitem[{Jia et~al.(2021)Jia, Pramanik, Saha~Roy, and Weikum}]{jia2021complex}
Zhen Jia, Soumajit Pramanik, Rishiraj Saha~Roy, and Gerhard Weikum. 2021.
\newblock Complex temporal question answering on knowledge graphs.
\newblock In \emph{Proceedings of the 30th ACM International Conference on Information \& Knowledge Management}, pages 792--802.

\bibitem[{Kannen et~al.(2022)Kannen, Sharma, Neelam, Khandelwal, Ikbal, Karanam, and Subramaniam}]{kannen2022targeted}
Nithish Kannen, Udit Sharma, Sumit Neelam, Dinesh Khandelwal, Shajith Ikbal, Hima Karanam, and L~Venkata Subramaniam. 2022.
\newblock Targeted extraction of temporal facts from textual resources for improved temporal question answering over knowledge bases.
\newblock \emph{arXiv preprint arXiv:2203.11054}.

\bibitem[{Krishnamurthy et~al.(2017)Krishnamurthy, Dasigi, and Gardner}]{krishnamurthy2017neural}
Jayant Krishnamurthy, Pradeep Dasigi, and Matt Gardner. 2017.
\newblock \href {https://doi.org/10.18653/v1/D17-1160} {Neural semantic parsing with type constraints for semi-structured tables}.
\newblock In \emph{Proceedings of the 2017 Conference on Empirical Methods in Natural Language Processing}, pages 1516--1526, Copenhagen, Denmark. Association for Computational Linguistics.

\bibitem[{Li et~al.(2021)Li, Fang, Lou, and Li}]{li-etal-2021-twt-table}
Tongliang Li, Lei Fang, Jian-Guang Lou, and Zhoujun Li. 2021.
\newblock \href {https://doi.org/10.18653/v1/2021.findings-emnlp.107} {{TWT}: Table with written text for controlled data-to-text generation}.
\newblock In \emph{Findings of the Association for Computational Linguistics: EMNLP 2021}, pages 1244--1254, Punta Cana, Dominican Republic. Association for Computational Linguistics.

\bibitem[{Lin et~al.(2020)Lin, Socher, and Xiong}]{lin2020bridging}
Xi~Victoria Lin, Richard Socher, and Caiming Xiong. 2020.
\newblock \href {https://doi.org/10.18653/v1/2020.findings-emnlp.438} {Bridging textual and tabular data for cross-domain text-to-{SQL} semantic parsing}.
\newblock In \emph{Findings of the Association for Computational Linguistics: EMNLP 2020}, pages 4870--4888, Online. Association for Computational Linguistics.

\bibitem[{Lyu et~al.(2023)Lyu, Havaldar, Stein, Zhang, Rao, Wong, Apidianaki, and Callison-Burch}]{lyu-etal-2023-faithful}
Qing Lyu, Shreya Havaldar, Adam Stein, Li~Zhang, Delip Rao, Eric Wong, Marianna Apidianaki, and Chris Callison-Burch. 2023.
\newblock \href {https://doi.org/10.18653/v1/2023.ijcnlp-main.20} {Faithful chain-of-thought reasoning}.
\newblock In \emph{Proceedings of the 13th International Joint Conference on Natural Language Processing and the 3rd Conference of the Asia-Pacific Chapter of the Association for Computational Linguistics (Volume 1: Long Papers)}, pages 305--329, Nusa Dua, Bali. Association for Computational Linguistics.

\bibitem[{Mavromatis et~al.(2021)Mavromatis, Subramanyam, Ioannidis, Adeshina, Howard, Grinberg, Hakim, and Karypis}]{mavromatis2021tempoqr}
Costas Mavromatis, Prasanna~Lakkur Subramanyam, Vassilis~N. Ioannidis, Soji Adeshina, Phillip~R. Howard, Tetiana Grinberg, Nagib Hakim, and George Karypis. 2021.
\newblock \href {http://arxiv.org/abs/2112.05785} {Tempoqr: Temporal question reasoning over knowledge graphs}.

\bibitem[{Nan et~al.(2021)Nan, Radev, Zhang, Rau, Sivaprasad, Hsieh, Tang, Vyas, Verma, Krishna, Liu, Irwanto, Pan, Rahman, Zaidi, Mutuma, Tarabar, Gupta, Yu, Tan, Lin, Xiong, Socher, and Rajani}]{radev2020dart}
Linyong Nan, Dragomir Radev, Rui Zhang, Amrit Rau, Abhinand Sivaprasad, Chiachun Hsieh, Xiangru Tang, Aadit Vyas, Neha Verma, Pranav Krishna, Yangxiaokang Liu, Nadia Irwanto, Jessica Pan, Faiaz Rahman, Ahmad Zaidi, Mutethia Mutuma, Yasin Tarabar, Ankit Gupta, Tao Yu, Yi~Chern Tan, Xi~Victoria Lin, Caiming Xiong, Richard Socher, and Nazneen~Fatema Rajani. 2021.
\newblock \href {https://doi.org/10.18653/v1/2021.naacl-main.37} {{DART}: Open-domain structured data record to text generation}.
\newblock In \emph{Proceedings of the 2021 Conference of the North American Chapter of the Association for Computational Linguistics: Human Language Technologies}, pages 432--447, Online. Association for Computational Linguistics.

\bibitem[{Neelam et~al.(2022)Neelam, Sharma, Karanam, Ikbal, Kapanipathi, Abdelaziz, Mihindukulasooriya, Lee, Srivastava, Pendus, Dana, Garg, Fokoue, Bhargav, Khandelwal, Ravishankar, Gurajada, Chang, Uceda-Sosa, Roukos, Gray, Lima, Riegel, Luus, and Subramaniam}]{neelam2022benchmark}
Sumit Neelam, Udit Sharma, Hima Karanam, Shajith Ikbal, Pavan Kapanipathi, Ibrahim Abdelaziz, Nandana Mihindukulasooriya, Young-Suk Lee, Santosh Srivastava, Cezar Pendus, Saswati Dana, Dinesh Garg, Achille Fokoue, G~P~Shrivatsa Bhargav, Dinesh Khandelwal, Srinivas Ravishankar, Sairam Gurajada, Maria Chang, Rosario Uceda-Sosa, Salim Roukos, Alexander Gray, Guilherme Lima, Ryan Riegel, Francois Luus, and L~Venkata Subramaniam. 2022.
\newblock \href {http://arxiv.org/abs/2201.05793} {A benchmark for generalizable and interpretable temporal question answering over knowledge bases}.

\bibitem[{Neeraja et~al.(2021)Neeraja, Gupta, and Srikumar}]{neeraja-etal-2021-incorporating}
J.~Neeraja, Vivek Gupta, and Vivek Srikumar. 2021.
\newblock \href {https://doi.org/10.18653/v1/2021.naacl-main.224} {Incorporating external knowledge to enhance tabular reasoning}.
\newblock In \emph{Proceedings of the 2021 Conference of the North American Chapter of the Association for Computational Linguistics: Human Language Technologies}, pages 2799--2809, Online. Association for Computational Linguistics.

\bibitem[{Ning et~al.(2020)Ning, Wu, Han, Peng, Gardner, and Roth}]{ning-etal-2020-torque}
Qiang Ning, Hao Wu, Rujun Han, Nanyun Peng, Matt Gardner, and Dan Roth. 2020.
\newblock \href {https://doi.org/10.18653/v1/2020.emnlp-main.88} {{TORQUE}: A reading comprehension dataset of temporal ordering questions}.
\newblock In \emph{Proceedings of the 2020 Conference on Empirical Methods in Natural Language Processing (EMNLP)}, pages 1158--1172, Online. Association for Computational Linguistics.

\bibitem[{Ning et~al.(2018)Ning, Zhou, Feng, Peng, and Roth}]{ning-etal-2018-cogcomptime}
Qiang Ning, Ben Zhou, Zhili Feng, Haoruo Peng, and Dan Roth. 2018.
\newblock \href {https://doi.org/10.18653/v1/D18-2013} {{C}og{C}omp{T}ime: A tool for understanding time in natural language}.
\newblock In \emph{Proceedings of the 2018 Conference on Empirical Methods in Natural Language Processing: System Demonstrations}, pages 72--77, Brussels, Belgium. Association for Computational Linguistics.

\bibitem[{Oguz et~al.(2020)Oguz, Chen, Karpukhin, Peshterliev, Okhonko, Schlichtkrull, Gupta, Mehdad, and Yih}]{oguz2020unified}
Barlas Oguz, Xilun Chen, Vladimir Karpukhin, Stan Peshterliev, Dmytro Okhonko, Michael Schlichtkrull, Sonal Gupta, Yashar Mehdad, and Scott Yih. 2020.
\newblock \href {"https://arxiv.org/abs/2012.14610"} {Unified open-domain question answering with structured and unstructured knowledge}.
\newblock \emph{arXiv preprint arXiv:2012.14610}.

\bibitem[{Parikh et~al.(2020)Parikh, Wang, Gehrmann, Faruqui, Dhingra, Yang, and Das}]{parikh2020totto}
Ankur Parikh, Xuezhi Wang, Sebastian Gehrmann, Manaal Faruqui, Bhuwan Dhingra, Diyi Yang, and Dipanjan Das. 2020.
\newblock \href {https://doi.org/10.18653/v1/2020.emnlp-main.89} {{ToTTo}: A controlled table-to-text generation dataset}.
\newblock In \emph{Proceedings of the 2020 Conference on Empirical Methods in Natural Language Processing (EMNLP)}, pages 1173--1186, Online. Association for Computational Linguistics.

\bibitem[{Pasupat and Liang(2015)}]{pasupat2015compositional}
Panupong Pasupat and Percy Liang. 2015.
\newblock \href {https://doi.org/10.3115/v1/P15-1142} {Compositional semantic parsing on semi-structured tables}.
\newblock In \emph{Proceedings of the 53rd Annual Meeting of the Association for Computational Linguistics and the 7th International Joint Conference on Natural Language Processing (Volume 1: Long Papers)}, pages 1470--1480, Beijing, China. Association for Computational Linguistics.

\bibitem[{Pramanick and Bhattacharya(2021)}]{pramanick-bhattacharya-2021-joint}
Aniket Pramanick and Indrajit Bhattacharya. 2021.
\newblock \href {https://doi.org/10.18653/v1/2021.eacl-main.102} {Joint learning of representations for web-tables, entities and types using graph convolutional network}.
\newblock In \emph{Proceedings of the 16th Conference of the European Chapter of the Association for Computational Linguistics: Main Volume}, pages 1197--1206, Online. Association for Computational Linguistics.

\bibitem[{Saxena et~al.(2021)Saxena, Chakrabarti, and Talukdar}]{saxena-etal-2021-question}
Apoorv Saxena, Soumen Chakrabarti, and Partha Talukdar. 2021.
\newblock \href {https://doi.org/10.18653/v1/2021.acl-long.520} {Question answering over temporal knowledge graphs}.
\newblock In \emph{Proceedings of the 59th Annual Meeting of the Association for Computational Linguistics and the 11th International Joint Conference on Natural Language Processing (Volume 1: Long Papers)}, pages 6663--6676, Online. Association for Computational Linguistics.

\bibitem[{Shang et~al.(2021)Shang, Qi, Wang, Huang, Wu, and Zhou}]{shang2021open}
Chao Shang, Peng Qi, Guangtao Wang, Jing Huang, Youzheng Wu, and Bowen Zhou. 2021.
\newblock \href {https://openreview.net/forum?id=li-3nHhT0xc} {Open temporal relation extraction for question answering}.
\newblock In \emph{3rd Conference on Automated Knowledge Base Construction}.

\bibitem[{Srivastava et~al.(2023)Srivastava, Rastogi, Rao, Shoeb, and Abid}]{srivastava2023imitation}
Aarohi Srivastava, Abhinav Rastogi, Abhishek Rao, Abu Awal~Md Shoeb, and Abubakar Abid. 2023.
\newblock \href {http://arxiv.org/abs/2206.04615} {Beyond the imitation game: Quantifying and extrapolating the capabilities of language models}.

\bibitem[{Srivastava et~al.(2024)Srivastava, Malik, Gupta, Ganu, and Roth}]{srivastava2024evaluating}
Pragya Srivastava, Manuj Malik, Vivek Gupta, Tanuja Ganu, and Dan Roth. 2024.
\newblock \href {http://arxiv.org/abs/2402.11194} {Evaluating llms' mathematical reasoning in financial document question answering}.

\bibitem[{Sui et~al.(2024)Sui, Zhou, Zhou, Han, and Zhang}]{sui2024table}
Yuan Sui, Mengyu Zhou, Mingjie Zhou, Shi Han, and Dongmei Zhang. 2024.
\newblock Table meets llm: Can large language models understand structured table data? a benchmark and empirical study.
\newblock In \emph{Proceedings of the 17th ACM International Conference on Web Search and Data Mining}, pages 645--654.

\bibitem[{Sun et~al.(2016)Sun, Ma, He, Yih, Su, and Yan}]{sun2016table}
Huan Sun, Hao Ma, Xiaodong He, Scott Wen-tau Yih, Yu~Su, and Xifeng Yan. 2016.
\newblock \href {https://www.microsoft.com/en-us/research/publication/table-cell-search-for-question-answering/} {Table cell search for question answering}.
\newblock In \emph{Proceedings of the companion publication of the 25th international conference on World Wide Web}. ACM - Association for Computing Machinery.

\bibitem[{Wang and Zhao(2024)}]{wang2024tram}
Yuqing Wang and Yun Zhao. 2024.
\newblock \href {http://arxiv.org/abs/2310.00835} {Tram: Benchmarking temporal reasoning for large language models}.

\bibitem[{Wei et~al.(2023)Wei, Wang, Schuurmans, Bosma, Ichter, Xia, Chi, Le, and Zhou}]{wei2023chainofthought}
Jason Wei, Xuezhi Wang, Dale Schuurmans, Maarten Bosma, Brian Ichter, Fei Xia, Ed~Chi, Quoc Le, and Denny Zhou. 2023.
\newblock \href {http://arxiv.org/abs/2201.11903} {Chain-of-thought prompting elicits reasoning in large language models}.

\bibitem[{Wen et~al.(2021)Wen, Qu, Ji, Ning, Han, Sil, Tong, and Roth}]{wen-etal-2021-event}
Haoyang Wen, Yanru Qu, Heng Ji, Qiang Ning, Jiawei Han, Avi Sil, Hanghang Tong, and Dan Roth. 2021.
\newblock \href {https://doi.org/10.18653/v1/2021.naacl-main.6} {Event time extraction and propagation via graph attention networks}.
\newblock In \emph{Proceedings of the 2021 Conference of the North American Chapter of the Association for Computational Linguistics: Human Language Technologies}, pages 62--73, Online. Association for Computational Linguistics.

\bibitem[{Yin et~al.(2020)Yin, Neubig, tau Yih, and Riedel}]{yin2020tabert}
Pengcheng Yin, Graham Neubig, Wen tau Yih, and Sebastian Riedel. 2020.
\newblock \href {http://arxiv.org/abs/2005.08314} {Tabert: Pretraining for joint understanding of textual and tabular data}.

\bibitem[{Yoran et~al.(2021)Yoran, Talmor, and Berant}]{yoran2021turning}
Ori Yoran, Alon Talmor, and Jonathan Berant. 2021.
\newblock \href {https://arxiv.org/pdf/2107.07261.pdf} {Turning tables: Generating examples from semi-structured tables for endowing language models with reasoning skills}.
\newblock \emph{arXiv preprint arXiv:2107.07261. Version 1.}

\bibitem[{Yu et~al.(2018)Yu, Zhang, Yang, Yasunaga, Wang, Li, Ma, Li, Yao, Roman, Zhang, and Radev}]{yu-etal-2018-spider}
Tao Yu, Rui Zhang, Kai Yang, Michihiro Yasunaga, Dongxu Wang, Zifan Li, James Ma, Irene Li, Qingning Yao, Shanelle Roman, Zilin Zhang, and Dragomir Radev. 2018.
\newblock \href {https://doi.org/10.18653/v1/D18-1425} {{S}pider: A large-scale human-labeled dataset for complex and cross-domain semantic parsing and text-to-{SQL} task}.
\newblock In \emph{Proceedings of the 2018 Conference on Empirical Methods in Natural Language Processing}, pages 3911--3921, Brussels, Belgium. Association for Computational Linguistics.

\bibitem[{Zayats et~al.(2021)Zayats, Toutanova, and Ostendorf}]{zayats2021representations}
Vicky Zayats, Kristina Toutanova, and Mari Ostendorf. 2021.
\newblock \href {https://doi.org/10.18653/v1/2021.eacl-main.253} {Representations for question answering from documents with tables and text}.
\newblock In \emph{Proceedings of the 16th Conference of the European Chapter of the Association for Computational Linguistics: Main Volume}, pages 2895--2906, Online. Association for Computational Linguistics.

\bibitem[{Zhang et~al.(2020{\natexlab{a}})Zhang, Wang, Wang, Cao, Zhang, and Wang}]{zhang-etal-2020-table}
Hongzhi Zhang, Yingyao Wang, Sirui Wang, Xuezhi Cao, Fuzheng Zhang, and Zhongyuan Wang. 2020{\natexlab{a}}.
\newblock \href {https://doi.org/10.18653/v1/2020.emnlp-main.126} {Table fact verification with structure-aware transformer}.
\newblock In \emph{Proceedings of the 2020 Conference on Empirical Methods in Natural Language Processing (EMNLP)}, pages 1624--1629, Online. Association for Computational Linguistics.

\bibitem[{Zhang et~al.(2019)Zhang, Zhang, and Balog}]{Zhang_2019}
Li~Zhang, Shuo Zhang, and Krisztian Balog. 2019.
\newblock \href {https://doi.org/10.1145/3331184.3331333} {Table2vec: Neural word and entity embeddings for table population and retrieval}.
\newblock In \emph{Proceedings of the 42nd International ACM SIGIR Conference on Research and Development in Information Retrieval}, SIGIR ’19. ACM.

\bibitem[{Zhang and Choi(2021)}]{zhang-choi-2021-situatedqa}
Michael Zhang and Eunsol Choi. 2021.
\newblock \href {https://doi.org/10.18653/v1/2021.emnlp-main.586} {{S}ituated{QA}: Incorporating extra-linguistic contexts into {QA}}.
\newblock In \emph{Proceedings of the 2021 Conference on Empirical Methods in Natural Language Processing}, pages 7371--7387, Online and Punta Cana, Dominican Republic. Association for Computational Linguistics.

\bibitem[{Zhang and Balog(2019)}]{Zhang:2019:ADC}
Shuo Zhang and Krisztian Balog. 2019.
\newblock \href {https://doi.org/10.1145/3357384.3357932} {Auto-completion for data cells in relational tables}.
\newblock In \emph{Proceedings of the 28th ACM International Conference on Information and Knowledge Management}, CIKM '19, pages 761--770, New York, NY, USA. ACM.

\bibitem[{Zhang and Balog(2020)}]{Zhang:2020:WTE}
Shuo Zhang and Krisztian Balog. 2020.
\newblock \href {https://doi.org/10.1145/3372117} {Web table extraction, retrieval, and augmentation: A survey}.
\newblock \emph{ACM Trans. Intell. Syst. Technol.}, 11(2):13:1--13:35.

\bibitem[{Zhang et~al.(2020{\natexlab{b}})Zhang, Dai, Balog, and Callan}]{Zhang:2020:SET}
Shuo Zhang, Zhuyun Dai, Krisztian Balog, and Jamie Callan. 2020{\natexlab{b}}.
\newblock \href {https://doi.org/10.1145/3397271.3401205} {Summarizing and exploring tabular data in conversational search}.
\newblock In \emph{Proceedings of the 43rd International ACM SIGIR Conference on Research and Development in Information Retrieval}, SIGIR '20, pages 1537--1540, New York, NY, USA. Association for Computing Machinery.

\end{thebibliography}
\bibliographystyle{acl_natbib}

\appendix
\section{Appendix}
\label{sec:appendix}

\begin{table*}[!htp]
\setlength{\tabcolsep}{2.0pt}
\centering
\scriptsize
\begin{tabular}{>{\raggedright}p{2.5cm}|>{\raggedright\arraybackslash}p{13.5cm}}
\hline
\textbf{Ordering} & \textbf{Q}: Arrange the following events in chronological order: (1) Brusilov Offensive by Russia. (2) Kamehameha I of the Island of Hawaii defeats the Oahuans at the Battle of Nu'uanu. (3) The Kuomintang, the Chinese nationalist party, is founded. (4) Emperor Claudius dies and is succeeded by his grand nephew Nero. (5) St. Norbert and 29 companions make their solemn vows marking the beginning of the Premonstratensian Order. \\
(Facts) & \textbf{A}. (1), (2), (4), (5), (3) \textcolor{red}{$\times$} \quad \textbf{B}. (4), (5), (2), (3), (1) \textbf{\cmark} \quad \textbf{C}. (3), (1), (2), (4), (5) \textcolor{red}{$\times$} \\
\hline
\textbf{Frequency}  & \textbf{Q}: It is also a love story, between Ace and Tobio, a trans woman. How often do they break up? \\
(Commonsense) & \textbf{A}. Once \textbf{\cmark} \quad \textbf{B}. Always \textcolor{red}{$\times$} \quad \textbf{C}. Once per week \textcolor{red}{$\times$} \\
\hline
\textbf{Duration} & \textbf{Q}: While Yoga Session gave attendees time to plant an entire garden, Jazz Concert was enough to water a few plants, and Board Game Night was merely smelling a flower. Which event was the longest? \\
(Analogy Inference) & \textbf{A}. Jazz Concert \textcolor{red}{$\times$} \quad \textbf{B}. Board Game Night \textcolor{red}{$\times$} \quad \textbf{C}. Yoga Session \textbf{\cmark} \\
\hline
\textbf{Typical Time}  & \textbf{Q}: Which event typically happens earlier: morning yoga or farmer starting their day? \\ 
(Comparison) & \textbf{A}. Morning yoga \textcolor{red}{$\times$} \quad \textbf{B}. Farmer starting their day \textbf{\cmark} \quad\textbf{C}.Around the same time\textcolor{red}{$\times$} \\
\hline
\textbf{Ambiguity Resolution}  & \textbf{Q}: The dynasty which fell in 1830 had risen to power roughly 90 years earlier. When was its establishment? \\ 
(Long-term Shift) & \textbf{A}.  1742 \textcolor{red}{$\times$} \quad \textbf{B}.  1745 \textcolor{red}{$\times$} \quad\textbf{C} 1740 \textbf{\cmark} \\
\hline
\textbf{Arithmetic}  & \textbf{Q}: In which week of year 2007 does the date 10-12-2007 occur? \\ 
(Week Identification) & \textbf{A}.Week41 \textbf{\cmark} \quad \textbf{B}.Week28 \textcolor{red}{$\times$} \quad\textbf{C}.Week5 \textcolor{red}{$\times$} \quad\textbf{D}.Week10 \textcolor{red}{$\times$} \\
\hline
\textbf{Temporal Relation}  & \textbf{Q}: It added that the Ministry of Economic Affairs and Finance was assigned to draw up practical procedure for
the ceding, while the Ministry of Welfare and Social Security would be responsible for identifying the
beneficiaries in two months. What is the relationship between the event ‘added’ and the event ‘ceding’? \\ 
  & \textbf{A}. IS\_INCLUDED \textcolor{red}{$\times$} \quad \textbf{B}. SIMULTANEOUS \textcolor{red}{$\times$} \quad\textbf{C}. AFTER \textbf{\cmark} \\
\hline
\textbf{Temporal NLI}  & \textbf{Q}: Premise: Two guys playing football on a campus green. 
 Hypothesis: They are practicing before the big game tomorrow \\ 
 & \textbf{A}. Entailment \textcolor{red}{$\times$} \quad \textbf{B}. Neutral \textbf{\cmark} \quad\textbf{C}. Contradiction\textcolor{red}{$\times$} \\
\hline
\textbf{Temporal Causality}  & \textbf{Q}:  The seasons changed from summer to autumn. What's the more plausible RESULT? \\ 
(Effect) & \textbf{A}. People evacuated their homes. \textcolor{red}{$\times$} \quad \textbf{B}. Leaves fell from the trees. \textbf{\cmark} \\
\hline
\textbf{Temporal Storytelling}  & \textbf{Q}: There is a huge clock in my living room. I turned the clock back one hour for daylight savings. My wife
also turned the clock back one hour for daylight savings. Our 2 kids each turned the clock back one hour for daylight savings. Which of the two endings is the most plausible correct ending to the story? \\ 
 & \textbf{A}. Then we wondered why it got so dark so early. \textbf{\cmark} \quad \textbf{B}. The kids were not happy \textcolor{red}{$\times$} \\
 \hline
\end{tabular}
\caption{\small Examples Questions in TRAM Dataset.}
\label{fig: tram_examples}
\end{table*}

\begin{table*}[h]
\centering 
\scriptsize
\setlength{\tabcolsep}{2.0pt}
\vspace{1cm}

\begin{tabular}{>{\raggedright}p{1.5cm}|>{\raggedright\arraybackslash}p{1cm}|>{\raggedright\arraybackslash}p{7cm}|>{\raggedright\arraybackslash}p{1cm}|>{\raggedright\arraybackslash}p{2cm}|>{\raggedright\arraybackslash}p{2cm}}
    \toprule
    \textbf{Task} & \textbf{Data Size} & \textbf{$\#$ Problem Types} & \textbf{Metrics} & \textbf{Answer Type} & \textbf{Text Sources} \\ 
    \midrule
    \multicolumn{6}{c}{Foundational Temporal Understanding Tasks} \\
    \midrule
    Ordering & 29,462 & Commonsense, Facts & Acc. &  3-Way MC & MCTACO$^{1}$, Wikipedia, Misc.\\
    \hline
    Frequency & 4,658 & Commonsense, Reading Comprehension, Application, Computation, Comparison, Facts & Acc. & 3-Way MC & MCTACO$^{1}$, SQuAD$^{2}$, Misc. \\
    \hline
    Duration & 7,232 & Commonsense, Reading Comprehension, Analogy Inference, Computation, Direct Comparison, Multi-step Comparison, Facts & Acc. & 3-Way MC & Same\\
    \hline
    Typical Time & 13,018 & Commonsense, Comparison, Facts, Reading Comprehension & Acc.& 3-Way MC & Same\\
    \midrule
    \multicolumn{6}{c}{Temporal Interpretation and Computation Tasks} \\
    \midrule
    Amb. Res. & 3,649 & Interpretation, Calendar shift, Long-term shift, Mid-term shift, Short-term shift & Acc. & 3-Way MC & Misc. \\
    \hline
    Arithmetic & 15,629 & Application, Date Computation, 12-hour Adjustment, 24-hour Adjustment, Month Shift, Week Identification, Year Shift, Time Computation, Time Zone Conversion & Acc. & 4-Way MC & Same \\
    \midrule
    \multicolumn{6}{c}{Advanced Temporal and Conceptual Understanding Tasks} \\
    \midrule
    Relation & 102,462 & - & Acc./F1 & 3-Way MC & TempEval-3$^{3}$\\
    \hline
    Temporal NLI & 282,144 & - & Acc./F1 & 3-Way MC & MNLI$^{4}$, SNLI$^{5}$\\
    \hline
    Causality & 1,200 & Cause, Effect & Acc. & 2-Way MC & COPA$^{6}$, Misc.\\
    \hline
    Storytelling & 67,214 & - & Acc. & 2-Way MC & ROC$^{7}$, SCT$^{8}$ \\
    \bottomrule
\end{tabular}

\caption{\small Overview of tasks in TRAM.}
\label{tab: tram_overview}
\end{table*}

\begin{figure*}[htp]
    \centering
    \includegraphics[width=0.9\textwidth]{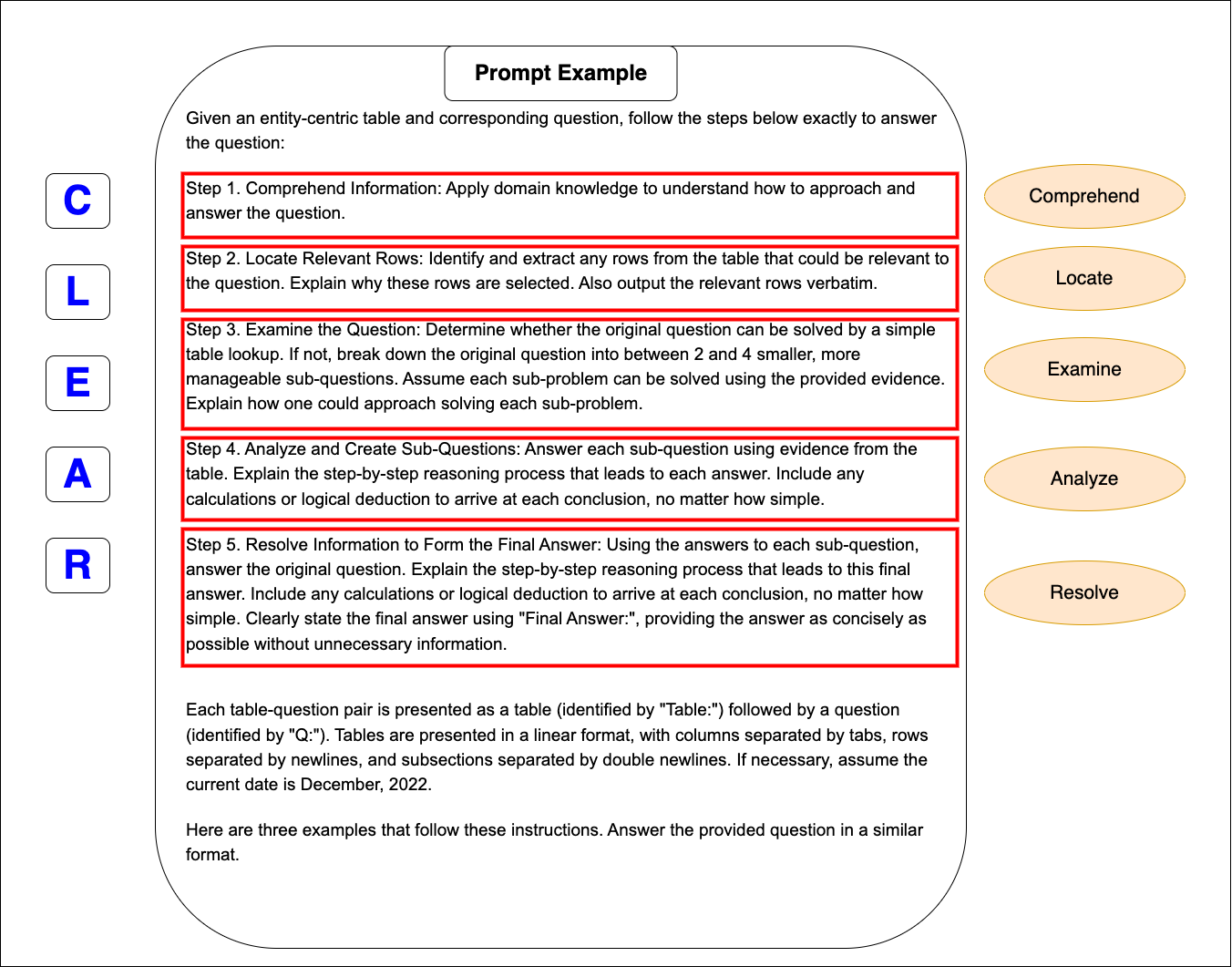} 
    \caption{\small Prompt Example}
    \label{fig:prompt_example}
\end{figure*}

\begin{figure*}[htp]
    \centering
    \includegraphics[width=0.9\textwidth]{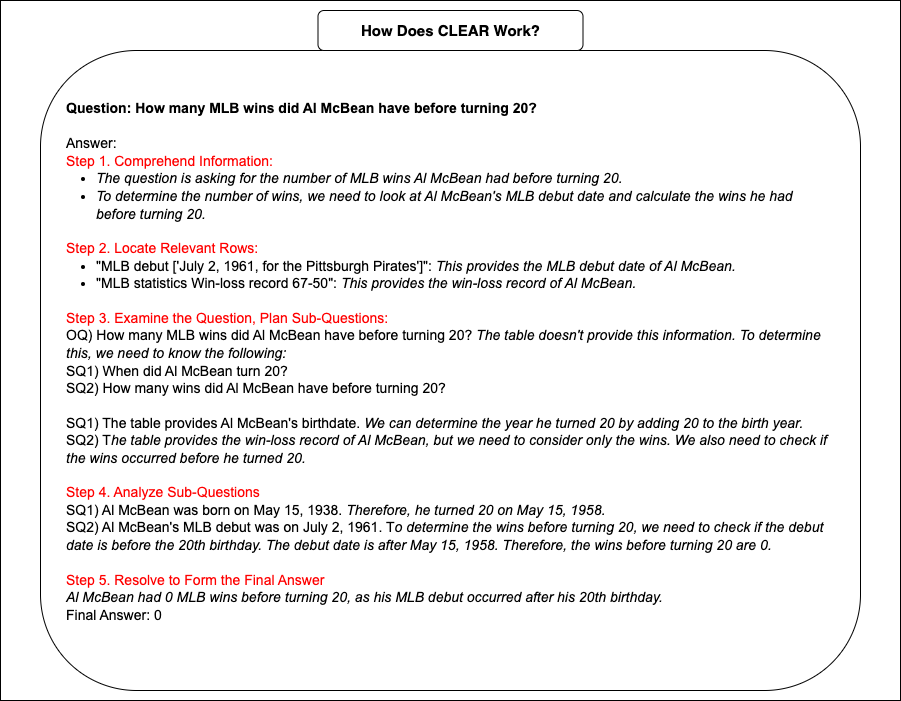} 
    \caption{\small The figure illustrates the step-by-step process of C.L.E.A.R instruction. The reference table is provided in Figure \ref{fig:wiki_table}}
    \label{fig:clear_working}
\end{figure*}

\begin{figure*}[!htb]
\centering
\begin{tcolorbox}[colback=white,arc=1mm] 
\begin{minipage}{\dimexpr\textwidth-2\fboxsep-2\fboxrule}
\centering 
\raggedright
\textbf{Input :} \\

\vspace{\baselineskip}
\small
Given an entity-centric table and corresponding question, follow the steps below exactly to answer the question:

\textbf{Step 1.} Comprehend Information: Apply domain knowledge to understand how to approach and answer the question.

\textbf{Step 2.} Locate Relevant Rows: Identify and extract any rows from the table that could be relevant to the question. Explain why these rows are selected. Also output the relevant rows verbatim.

\textbf{Step 3.} Examine the Question: Determine whether the original question can be solved by a simple table lookup. If not, break down the original question into between 2 and 4 smaller, more manageable sub-questions. Assume each sub-problem can be solved using the provided evidence. Explain how one could approach solving each sub-problem.

\textbf{Step 4.} Analyze Sub-Questions: Answer each sub-question using evidence from the table. Explain the step-by-step reasoning process that leads to each answer. Include any calculations or logical deduction to arrive at each conclusion, no matter how simple.

\textbf{Step 5.} Resolve to Form the Final Answer: Using the answers to each sub-question, answer the original question. Explain the step-by-step reasoning process that leads to this final answer. Include any calculations or logical deduction to arrive at each conclusion, no matter how simple. Clearly state the final answer using "Final Answer:", providing the answer as concisely as possible without unnecessary information.\\[1\baselineskip]

Each table-question pair is presented as a table (identified by "Table:") followed by a question (identified by "Q:"). Tables are presented in a linear format, with columns separated by tabs, rows separated by newlines, and subsections separated by double newlines. If necessary, assume the current date is December, 2022.

========================

\textbf{Table:}

Title   Bette Davis \\
Died   \quad October 6, 1989 | (1989-10-06) | (aged 81) |\\ Neuilly-sur-Seine, France\\
Resting place   \quad Forest Lawn Memorial Park\\
Occupation      \quad Actress\\
Years active    \quad 1929-1989\\
Political party Democratic\\
Spouse(s)       \quad Harmon Oscar Nelson | ( | married |  1932; | divorced |  1938) | Arthur Farnsworth | ( | married |  1940; died 1943) | William Grant Sherry | ( | married |  1945; | divorced |  1950) | Gary Merrill | ( | married |  1950; | divorced |  1960) \\
Children        \quad 3, including B. D. Hyman \\
Bette Davis     \quad ['BETTEDavis (cropped).jpg | Publicity photo, 1935']\\[1\baselineskip]

\textbf{Question:} How many years after Davis's career began did she take the photo?

\end{minipage}
\end{tcolorbox}
\caption{Example 1 from TempTabQA head set with C.L.E.A.R prompting on GPT 3.5 turbo : Input}
\label{fig:example1:input}
\end{figure*}

\begin{figure*}[!htb]
\centering
\begin{tcolorbox}[colback=white,arc=1mm] 
\begin{minipage}{\dimexpr\textwidth-2\fboxsep-2\fboxrule}
\centering
\raggedright

\textbf{Response :} \\

\vspace{\baselineskip}
\small
\textbf{Step 1.} Comprehend Information
Domain Knowledge: \\
- The question asks "how many years", so the answer should be a length of time in years \\
- To determine the time between two years, take the absolute value of the difference between the two years.\\[1\baselineskip]

\textbf{Step 2.} Locate Relevant Rows\\
- "Title        Bette Davis": This indicates that the information in the table is about Bette Davis.
- "Years active 1929-1989": This provides the start and end year of Bette Davis's career.
- "Bette Davis  ['BETTEDavis (cropped).jpg | Publicity photo, 1935']": This provides the year the photo was taken.\\[1\baselineskip]
\textbf{Step 3.} Examine the Question
OQ) How many years after Davis's career began did she take the photo? The table doesn't provide this information. To determine the time between these two events, we need to know the following:
\begin{itemize}
    \item SQ1) When did Davis's career begin?
    \item SQ2) When did Davis take the photo? : 
 Davis took the photo in 1935
\end{itemize}

SQ1) When did Davis's career begin? \\The table gives the range years spanning Davis's career. We can determine the begin year of her career by looking at the start year of this range.

SQ2) When did Davis take the photo? \\The table provides the year the photo was taken.\\[1\baselineskip]

\textbf{Step 4.} Answer Sub-Questions
SQ1) Davis's career lasted from 1929 to 1989. Therefore, her career began in 1929.
SQ2) The photo was taken in 1935.
\\[1\baselineskip]
\textbf{Step 5.} Synthesize to Form the Final Answer
OQ) How many years after Davis's career began did she take the photo?
Davis's career began in 1929 \\(SQ1), and she took the photo in 1935 \\(SQ2). To get the number of years between these two events, take the absolute value of the difference between the two years. \\The absolute value of (1929-1935) is 6.
\\Therefore, the photo was taken 6 years after Davis's career began.\\[1\baselineskip]

\textbf{Final Answer: 6}

\end{minipage}
\end{tcolorbox}
\caption{Example 1 from TempTabQA head set with C.L.E.A.R prompting on GPT 3.5 turbo : Response}
\label{fig: example1_response}
\end{figure*}

\begin{figure*}[!htb]
\centering
\begin{tcolorbox}[colback=white,arc=1mm] 
\begin{minipage}{\dimexpr\textwidth-2\fboxsep-2\fboxrule}
\centering
\raggedright
\textbf{Input :} \\

\vspace{\baselineskip}
\small
Given an entity-centric table and corresponding question, follow the steps below exactly to answer the question:

\textbf{Step 1.} Comprehend Information: Apply domain knowledge to understand how to approach and answer the question.

\textbf{Step 2.} Locate Relevant Rows: Identify and extract any rows from the table that could be relevant to the question. Explain why these rows are selected. Also output the relevant rows verbatim.

\textbf{Step 3.} Examine the Question: Determine whether the original question can be solved by a simple table lookup. If not, break down the original question into between 2 and 4 smaller, more manageable sub-questions. Assume each sub-problem can be solved using the provided evidence. Explain how one could approach solving each sub-problem.

\textbf{Step 4.} Analyze Sub-Questions: Answer each sub-question using evidence from the table. Explain the step-by-step reasoning process that leads to each answer. Include any calculations or logical deduction to arrive at each conclusion, no matter how simple.

\textbf{Step 5.} Resolve to Form the Final Answer: Using the answers to each sub-question, answer the original question. Explain the step-by-step reasoning process that leads to this final answer. Include any calculations or logical deduction to arrive at each conclusion, no matter how simple. Clearly state the final answer using "Final Answer:", providing the answer as concisely as possible without unnecessary information.\\[1\baselineskip]

Each table-question pair is presented as a table (identified by "Table:") followed by a question (identified by "Q:"). Tables are presented in a linear format, with columns separated by tabs, rows separated by newlines, and subsections separated by double newlines. If necessary, assume the current date is December, 2022.

========================

\textbf{Table:}

Title   \quad Dwight Bernard\\
Dwight Bernard  \quad ['Dwight Bernard.jpg']\\

Pitcher\\

Born  \quad | ( | 1952-05-31 | ) | May 31, 1952 | (age 70) | Mount Vernon, Illinois\\

Batted \quad | Right | Threw | Right\\

MLB debut\\
MLB debut       \quad ['June 29, 1978, for the New York Mets']\\

Last MLB appearance\\
Last MLB appearance    \quad ['October 2, 1982, for the Milwaukee Brewers']\\

MLB statistics\\
Win-loss record \quad 4-8\\
Earned run average     \quad  4.14\\
Strikeouts      \quad 92\\

Teams\\
Teams  \quad ['New York Mets (1978-1979) | Milwaukee Brewers (1981-1982)']\\[1\baselineskip]

\textbf{Question:} In what year did Dwight Bernard start playing with the last team he was associated with?

\end{minipage}
\end{tcolorbox}
\caption{Example 2 from TempTabQA tail set with C.L.E.A.R prompting on GPT 3.5 turbo : Input}
\label{fig: example2_input}
\end{figure*}

\begin{figure*}[!htb]
\centering
\begin{tcolorbox}[colback=white,arc=1mm] 
\begin{minipage}{\dimexpr\textwidth-2\fboxsep-2\fboxrule}
\centering
\raggedright

\textbf{Response :} \\

\vspace{\baselineskip}
\small
\textbf{Step 1.} Comprehend Information\\
Domain Knowledge:\\
- The question asks "In what year", so the answer should be a year.\\
- To determine the last team someone was associated with, identify the one that occurred most recently.\\
- Given an event spanning a range of years, the start is the first number in that range.\\[1\baselineskip]

\textbf{Step 2.} Locate Relevant Rows\\
- "Title    \quad Dwight Bernard": This indicates that the information in the table is about Dwight Bernard.\\
- "Teams    \quad ['New York Mets (1978-1979) | Milwaukee Brewers (1981-1982)']": This provides a list of all teams Dwight Bernard played with and the years he played with each.\\[1\baselineskip]

\textbf{Step 3.} Examine the Question\\
OQ) In what year did Dwight Bernard start playing with the last team he was associated with? The table doesn't provide this information. To determine this, we need to know the following:\\
\begin{itemize}
    \item SQ1) What was the last team Dwight Bernard was associated with?
    \item SQ2) When did Dwight Bernard start playing for this team?
\end{itemize}
    
SQ1) What was the last team Dwight Bernard was associated with? \\The table doesn't provide this information. However, the table provides a list of all teams Bernard was associated with and the years he was associated with each team. We can determine the last team Dwight Bernard was associated with by comparing the last year he was associated with each of his teams.\\

SQ2) When did Dwight Bernard start playing for this team? \\The table provides the range of years Bernard was associated with each of his teams. We can determine the year he started playing for that team by looking at the start year in the range.\\[1\baselineskip]

\textbf{Step 4.} Analyze Sub-Questions\\
SQ1) The only teams Dwight Bernard has played with are the New York Mets and Milwaukee Brewers. Bernard played for the Mets from 1978 to 1979 and the Brewers from 1981 to 1982. He was associated with the Brewers in 1982, which is more recent than the Mets in 1979; therefore, the last team he was associated with is the Brewers.\\
SQ2) He played with the Brewers from 1981 to 1982, so he started playing for the Brewers in 1981.
\textbf{Step 5.} Resolve to Form the Final Answer\\
OQ) In what year did Dwight Bernard start playing with the last team he was associated with?\\
The last team Dwight Bernard was associated with was the Brewers (SQ1), which he started playing for in 1981 (SQ2).\\
Therefore, Dwight Bernard started playing with the last team he was associated with in 1981.\\[1\baselineskip]

\textbf{Final Answer: 1981}

\end{minipage}
\end{tcolorbox}
\caption{Example 2 from TempTabQA tail set with C.L.E.A.R prompting on GPT 3.5 turbo : Response}
\label{fig: example2_response}
\end{figure*}

\end{document}